RESEARCH ARTICLE

# Intelligent Sampling Consensus for Homography Estimation in Football Videos Using Featureless Unpaired Points

**GEORGE NOUSIAS[1], KONSTANTINOS K. DELIBASIS[1], AND ILIAS G. MAGLOGIANNIS[2], (Senior Member, IEEE)**

[1]Department of Computer Science and Biomedical Informatics, University of Thessaly, 351 31 Lamia, Greece
[2]Department of Digital Systems, University of Piraeus, 185 34 Athens, Greece

Corresponding author: Konstantinos K. Delibasis (kdelibasis@gmail.com)

This work was supported by European Union and Greek national funds through the Operational Program Competitiveness, Entrepreneurship and Innovation, through the Call RESEARCH—CREATE—INNOVATE (Project: DFVA—Deep Football Video Analytics) under Grant T2EKΔK-04581.

**ABSTRACT** Estimating the homography matrix between images captured under radically different camera poses and zoom factors is a complex challenge. Traditional methods rely on the Random Sample Consensus (RANSAC) algorithm, which requires pairs of homologous points, pre-matched based on local image feature vectors. Sampling consensus is a core step in many Artificial Intelligence (AI) algorithms that enable computer systems to recognize patterns in data. In this paper, we propose $H$-RANSAC, an algorithm for homography estimation that eliminates the need for feature vectors or explicit point pairing, while it optionally supports point labeling into two classes. $H$-RANSAC introduces a novel geometric (cheiral) criterion that intelligently rejects implausible point configurations at the beginning of each iteration, while leveraging concave quadrilaterals typically discarded by similar algorithms. A post-hoc criterion at the end of each iteration improves accuracy further. Analytical derivations of the expected maximum iterations are provided, considering success probabilities and outlier rates, enabling adaptive performance tuning. The algorithm is validated on a demanding task: estimating homography between video frames of football matches captured by 12 cameras with highly divergent viewpoints. Results show that $H$-RANSAC significantly outperforms state-of-the-art classical methods, combined with deep learning-based salient point detection, in terms of average reprojection error and success rates. The relevant implementation is available in https://github.com/gnousias/H-RANSAC

**INDEX TERMS** Featureless points, football video analytics, geometric constraints, homography estimation, intelligent sample consensus algorithm.

## I. INTRODUCTION

Recovering the homography matrix between two images is an important step in many image analysis problems. Homography transform grasps the geometric transform between two different central projections of the same planar points. Although a minimum of 4 point pairs are required to calculate homography, more image pairs are required for acceptable accuracy. RANSAC [1] is the most established family of algorithms for selecting point pairs to recover specific transformations. However, the classic RANSAC implementation available in OpenCV [2] library, or in the computer vision toolbox of Matlab [3], as well as most of its variants [3], [4], [5], [6], [7], [8], [9], [10], [11] require a set of possible point pairs, generated by a previous matching algorithm that is usually based on feature vector descriptors, such as Scale Invariant Feature Transform (SIFT), Speeded Up Robust Features (SURF), or more recently, deep learning methods [12], [13].

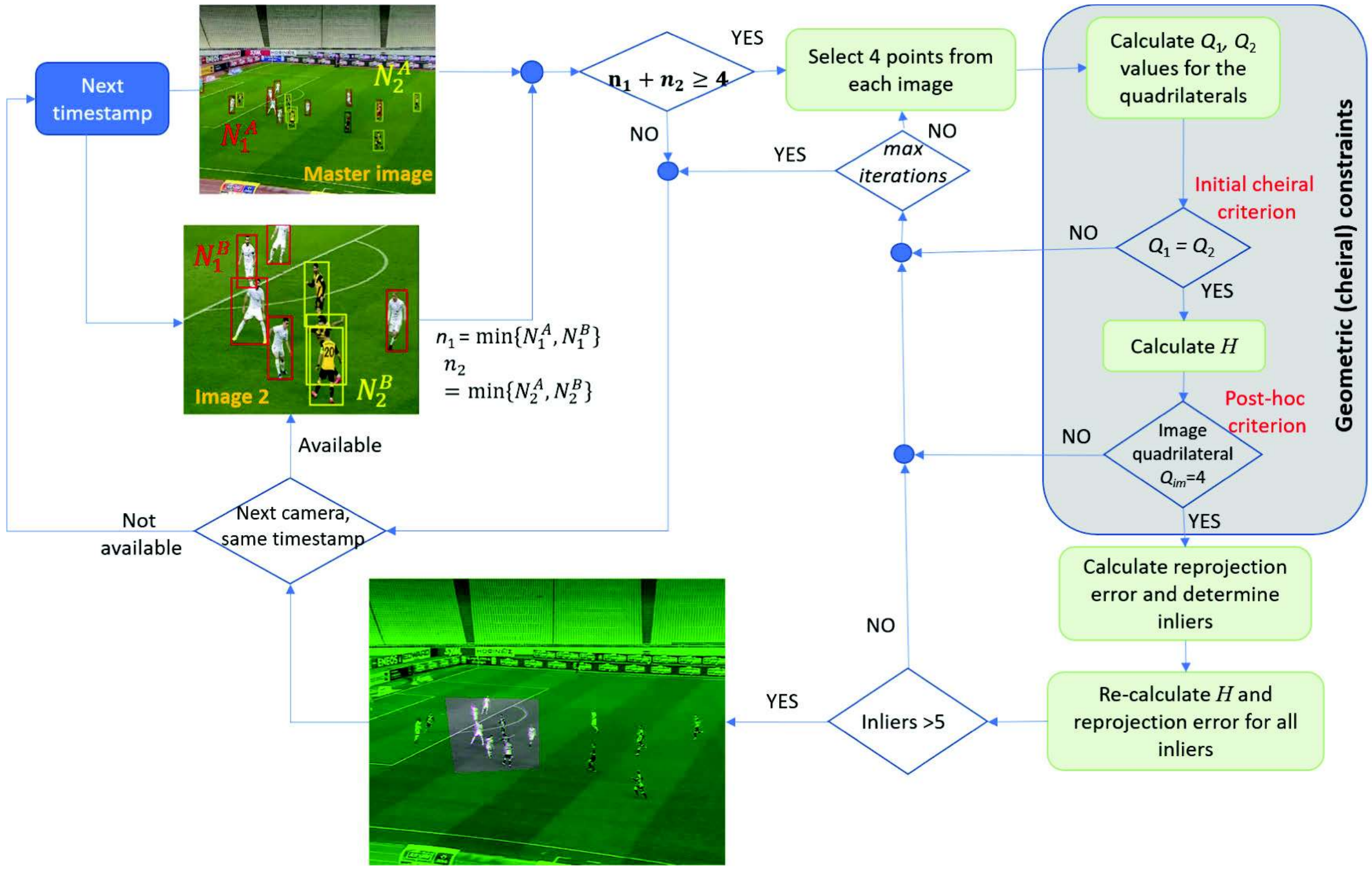

**FIGURE 1.** **A semantic workflow of the proposed methodology, for football images.**

In many applications a more generalized RANSAC implementation is required, that can handle 2 sets of featureless points. This is particularly applicable to analysis of football video sequences acquired from multiple cameras with radically different position, orientation and zoom factor, in order to extract analytics [14].

### A. RELATED WORK

Various approaches have been proposed for homography matrix estimation, using direct or feature-based methods. Direct methods, such as Lucas-Kanade [15] algorithm, optimize a cost function of pixel-to-pixel matching. Feature-based methods like SIFT [16], SURF [17] or ORB [18], combined with random sampling consensus algorithms (RANSAC) are more accurate and commonly-used.

A variety of algorithms of the RANSAC family have been proposed, to estimate model parameters in machine learning and computer vision, like Universal Framework for Random Sample Consensus (USAC) [4], VSAC [7], Progressive sample consensus (PROSAC) [5], Marginalizing Sample Consensus (MAGSAC++) [6], Graph-Cut [8], GroupSAC [9] or DEGENSAC [10], using various sampling, verification or optimization techniques for faster and often more accurate results. Torr and Zisserman proposed MLESAC [3], a new robust estimator with application to estimating image geometry, which is based on detected corner points, utilizing proximity and correlation information to form pairs of points. This method was further improved by Tordoff et al. resulting in Guided-MLESAC [11], a faster image transform estimation by using matching priors. However, both methods assume motion based images, using priors to refine the posterior probability of matches. Another approach that utilises motion between the two images to be aligned (“Bilateral functions”) was proposed in [19]. In Shi et al. [16], SIFT is combined with RANSAC with an intermediate step of removing non-plausible image pairs before invoking the RANSAC algorithm. In Hossein-nejad and Nasri [20] image registration is proposed based on SIFT features and RANSAC with adaptive threshold for inlier determination. Liu et al. [21] proposed homography recovering based on the straight lines captured in the moving and reference image, using the Canny edge detection and Hough transform. Line correspondences are estimated using a genetic algorithm [22]. However, although our application deals with football frames simultaneously acquired by multiple cameras, certain frames with high zoom factor lack any lines, rendering this approach not directly applicable. A different approach, “BB-Homography” [23], fuses fast binary descriptor matching and bipartite graph for homography estimation, however the authors report decreased accuracy as the view point angle increases. It should be noted that sport videos from multiple cameras often exhibit radically different viewpoints.

Deep learning has prevailed in almost any aspect of science [24], [25], thus more recent methodologies are estimating homography transformations using deep learning algorithms like GANs (Generative Adversarial Networks) or transformers. DeTone et al. [26], were first to propose a deep neural network with just 10 layers, producing an 8-DOF (Degree Of Freedom) homography. Nie et al. [27] proposed a contextual correlation layer for end-to-end multi-grid homography estimation network utilising depth information, enabling the alignment between images. In [28], an unsupervised learning algorithm is proposed where a deep convolutional neural network is trained to estimate planar homographies. A pixel-wise intensity error metric is minimized, without demanding ground truth, achieving same or better results than direct and feature-based methods.

In [12], DeTone et al. proposed "SuperPoint", a self-supervised framework where a fully convolutional model is trained to localize interest points and calculate its corresponding descriptors, performing equally well, or occasionally surpassing the classic feature detectors (SIFT, SURF or Oriented Fast and Rotated -ORB-). SuperGlue is an end-to-end trained graph neural network with attention [13], utilized to solve correspond point matching, using either classic descriptors or just a set of points from each image. SGA-Net is another deep learning approach [29] that inputs an image and returns a set of salient points with their local descriptors. Combining with any RANSAC variant, it generates corresponding point pairs between any two images and finally spatial image matching. Another similar approach [30] is SOSNet that returns image points with their learned local descriptors using second order similarity regularization. An approach that recovers corresponding point pairs from two different images that contain the same object using CUR matrix decomposition is described in [31]. In [32] a hardware architecture is proposed for accelerating classic image point detection and matching such as Features from Accelerated Segment Test (FAST) and Binary Robust Independent Elementary Features (BRIEF).

Hoang et al. [33] consider a deep learning approach for dynamic scenes rather than static images, that estimates and progressively refines homography, being able to cope with large global motion between the two images. In [34], an iterative homography network is considered. In Zhou et al. [35], Deep Homography Estimation is proposed, applied to wall mapping for wall-climbing robots, using center-aligned and non-aligned images. In [36], Shao et al. proposed a semantic filter to be utilized by RANSAC for Fundamental matrix estimation, using region-based convolutional neural networks. This approach produced robust results even in high percentage outliers. In [37], a modified RANSAC algorithm is proposed, namely LP-RANSAC. The algorithm adopts a locality preserving constraint, added in the original implementation of RANSAC, helping to discard most of the bad correspondences without significant computational cost. Also, a non-uniform sampling strategy based on the locality preserving scores is proposed without requiring any prior information. Zheng et al. [38] proposed an accelerated RANSAC variant, named PSSC-RANSAC (Prior Sampling & Sample Check RANSAC), utilizing image characteristics like contrast, coarseness and roughness in order to extract feature points.

A very recent approach [39] proposed the use of Geometrized Transformer (GeoFormer), a new detector-free feature matching method to generate pairs of matching points between two images. RANSAC variants can be used to subsequently recover homography transform between the images. MCNet, [40] an iterative deep homography estimation architecture, was proposed, based on Multiscale Correlation searching. Another work tackles the problem of homography estimation between images of different modalities and pixelation, using superresolution [41]. A recurrent framework for homography estimation was presented in [42]. DKM is a dense kernelized feature matching [43] that strives to discover all correspondences between two images, followed by refining steps to estimate homography transform.

It should be mentioned that certain deep learning-based approaches estimate the homography matrix indirectly, by outputting the displacements of the four image corners, usually in modest ranges [26]. However, in certain applications, such as the one discussed in the proposed work, the range of image corner displacements is far too radical for the reported deep learning methods. Furthermore, local feature vectors for image points may not be descriptive enough to establish pairs of points, or the radically different camera poses induce uncorrelated local image features, even for corresponding points. Such an example is shown from the dataset in our work in Fig. 1, where two images of a football stadium are acquired during a football match by a number of cameras. RANSAC implementations, such as the one available in OpenCV [2] library, or in the computer vision toolbox of Matlab [3] require a set of possible point pairs, based on corresponding feature vectors. Subsequently, RANSAC repeatedly selects randomly the necessary number of point pairs from the predetermined candidate pairs, until it discovers a plausible transform.

## II. METHODOLOGY

The proposed methodology generalizes the RANSAC algorithm so that it can operate on two different sets of unpaired image points, without any local feature vector to assign correspondence. This differentiates our method from the usual approaches that require feature-based matched points. In order to increase the versatility of the algorithm in the absence of predetermined point-pairs, facilitate the random point selection and decrease the necessary number of iterations, we modified RANSAC to utilize the assignment of the two sets of points into two different classes, although this is not mandatory. This is very convenient in the case of our application, which involves football, or any team-based sport. Additionally, a geometric constraint is tested before each iteration, based on the type of quadrilateral (convex and concave self-intersecting or non-self-intersecting) that is formed on each image by the four selected pairs of points. Finally, a post-hoc geometric constraint (at the end of each iteration) detects implausible transforms based on the aforementioned type of the transformed image quadrilateral. These constraints achieve a significant reduction of the required iterations. The use of geometric criteria follow the approach of introducing cheiral constraints in the standard RANSAC algorithm e.g., [4], [44], and [45]. However, our proposed constraints are more effective in utilizing selected point pairs, compared to the aforementioned methods. The main steps of the proposed algorithm are shown graphically in Fig. 1. In the next section we provide elaborate methodological comparisons between the proposed algorithm and the state-of-the-art methods. The proposed $H$-RANSAC has been applied to simultaneously acquired frames of a football game, outperforming other state-of-the-art methods in terms of reprojection error as well as in the number of successfully processed frame pairs. The proposed algorithm is applied to recover homography transform between video frames of football matches captured simultaneously by 12 cameras with highly divergent viewpoints and zoom factors. In order to apply this method to different types of datasets, it is prerequisite that each frame contains a flat surface with fiducial points on the surface. If the points are moving (such as the position of feet of walking/running people), then the different frames must be acquired simultaneously.

## A. NOTATION AND BACKGROUND

Homography transform between two images is defined by a $3 \times 3$ matrix, that has 8 degrees of freedom (DOF), or equivalently it is scalable. In many applications it is scaled by dividing all elements by $h_9$, or by normalizing the Frobenious norm:

$$H = \begin{bmatrix} h_1 & h_2 & h_3 \\ h_4 & h_5 & h_6 \\ h_7 & h_8 & h_9 \end{bmatrix}.$$

As it is well known, estimating $H$ for 4 pairs of corresponding points $p_i^A = (x_i^A, y_i^A) \in P^A, p_i^B = (x_i^B, y_i^B) \in P^B, i = 1, 2, 3, 4$ is calculated by solving a homogeneous system of linear equations, using the SVD method, resulting in a solution such that $\sum h_i^2 = 1$. The system can also be transformed into non-homogeneous by setting $h_9 = 1$, that can be solved using the least squares method. If more points pairs are available, then the system of equations becomes overdetermined and can still be solved similarly.

Although the proposed $H$-RANSAC operates on sets of points without predetermined point-pairs without requiring local feature vectors, it optionally utilizes assignment of the points up to two classes. It is important to note that the different classes of points co-exist in each of the two images. In the rest of the paper we assume that $p_{c,i}^A$ is the $i$th point of image $A$ that belongs to class $c$. Candidate valid image pairs (image pairs where a homography matrix may be calculated) are selected based on the number of points for each class available in each image. More specifically, let $N_1^A$,$N_2^A$ and $N_1^B$,$N_2^B$ be the number of players of $class_1$ and $class_2$ in images A and B, where $N^A = N_1^A + N_2^A, N^B = N_1^B + N_2^B$. The minimum number of points from each class, considering the two images are $n_1 = min(N_1^A, N_1^B)$ and $n_2 = min(N_2^A, N_2^B)$. Two images constitute a possibly valid image pair for Homography recovery only if at least 4 candidate point pairs exists. However, in our method where non previously matched points are handled, we enforce a stricter criterion:

$$n_1 + n_2 \geq 4. \tag{1}$$

In our application, the two classes of points correspond to the two teams of players, thus the term "class" and "team" may be used interchangeably.

## B. RANDOM POINT SELECTION USING TWO CLASSES OF POINTS

The proposed RANSAC algorithm is invoked to calculate homography matrix between two image frames, only if Eq.(1) holds. Homography estimation using RANSAC relies on random selection of sets of 4 points from each image. Since our algorithm does not require pairs of points (based on local feature vectors) for each image pair, we utilize the point assignment into two different classes, using biased roulette wheel selection [46]. More specifically, Algorithm 1 shows the steps for determining the class of each of the 4 point pairs, for the more general case of $n_c$ classes of points (in this application, $n_c = 2$).

**Algorithm 1** Determining the Class of Each Point Pair With Biased Roulette Wheel Selection

**Input:** $n_1, \ldots, n_{n_c}$
**Output:** $(class_1, \ldots, class_4)$ % the class of each of the 4 required point pairs
  $q_i \leftarrow \frac{n_i}{\sum_{j=1}^{n_c} n_j}, i = 1, \ldots, n_c$
  $s_i \leftarrow \sum_{k=1}^{i} q_k$
  **for** $k = 1$ to 4 **do**
    $\rho \leftarrow U(0, 1)$ % random value according to uniform distribution in [0,1)
    $class_k \leftarrow argmin(s_i > \rho)$

Let $e_1$ and $e_2$ be the numbers of selected points that belong to $class_1$ and $class_2$, respectively. Thus, $e_1 + e_2 = 4$ and $e_1 \leq n_1, e_2 \leq n_2$ and $e_1, e_2$ are analogous to the probability of class selection $e_1 \sim \frac{n_1}{n_1+n_2}$ and $e_2 \sim \frac{n_2}{n_1+n_2}$.

Thus, $e_1$ points are randomly selected from $N_1^A$ points (of class 1 in image A) and from $N_1^B$ points (of class 1 in image B). Similarly, $e_2$ points are randomly selected from $N_2^A$ points (of class 2 in image A) and from $N_2^B$ points (of class 2 image B). On each RANSAC iteration, $e_1 + e_2$ points are utilized to estimate the homography matrix.

If point assignment to classes is not available, then point selection is randomly performed, selecting one permutation of 4 points from each image.

## C. INLIER PAIR DETECTION AND CALCULATION OF H

Let $p_{c,i}^A, p_{c,i}^B$ for $0 < i \leq 4$ be the 4 randomly selected points from images A and B that belong to class $c$. The homography matrix $H$ that maps image B onto A is estimated based on these 4 selected pairs, as described above. Since there is no known correspondence between the two sets of points and since the points are featureless (except for the assignment to two classes), the following steps are performed to establish possible point pairs for the calculated $H$.

Every point of image B is reprojected using the recovered $H$. For every point $p^A_{c,i}$, the point $j$ of image B, $p^B_{c,j}$ is determined, such as the transformed point $Hp^B_{c,j}$ is the closest to $p^A_{c,i}$ and their distance is less than a threshold $T$, provided that the two points belong to the same class $c$. More formally $(i, j)$ is considered a pair candidate in the current iteration, referring to points of the same class $c$, if and only if

$$\|p^A_{c,i} - Hp^B_{c,j}\| = \min_k \|p^A_{c,i} - Hp^B_{c,k}\|, for\ k = 1, \ldots, N^B$$
$$AND\ \|p^A_{c,i} - Hp^B_{c,j}\| < T \quad (2)$$

The distance threshold is adaptively calculated for each image pair:

$$T = \lambda \times max\{\|p^B_i - p^B_j\|\}, 0 < i, j \leq N^B \quad (3)$$

where $\lambda$ is a parameter of the method with a typical value of 0.01. The effect of its value on the behavior of the proposed method is studied in the results section. Let $n_{inliers}$ be the number point pairs that satisfy Eq. 2. The aforementioned steps for homography estimation are depicted in Fig. 1. The algorithm is terminated if an $H$ with $n_{inliers} > 5$ is found, or if the maximum number of iterations $n_{iter}$ is reached.

### D. REFINING POINT SELECTION WITH GEOMETRIC CONSTRAINTS AT THE BEGINNING OF EACH ITERATION

The special case of homography from planar points allows for geometric testing and rejecting candidate point pairs before being used in the current iteration. The proposed algorithmic steps are formalized below.

*Definition 1:* Let us define a quadrilateral $\{p_1, p_2, p_3, p_4\}$ as an ordered set of 4 points.

*Definition 2:* A planar quadrilateral is convex if and only if all the angles $\theta_i$, $i = 1, \ldots, 4$ between consecutive edges are less than $\pi$. Otherwise, it is concave.

*Definition 3:* A planar concave quadrilateral is self-intersecting, if two of its line segments intersect (see Fig 2(b)).

A computational implementation of a convexity test according to the previous definition is the following. Let $p_i$ be the current point and $p_p$ and $p_n$ be the previous and next point in the ordered set of 4 points. Assuming that we are only interested in discriminating between angles less than or greater than $\pi$, then it is sufficient to calculate the sign of the z-coordinate of the cross product of the vectors of the two consecutive edges

$$v^i = (p_i - p_p) \times (p_n - p_i). \quad (4)$$

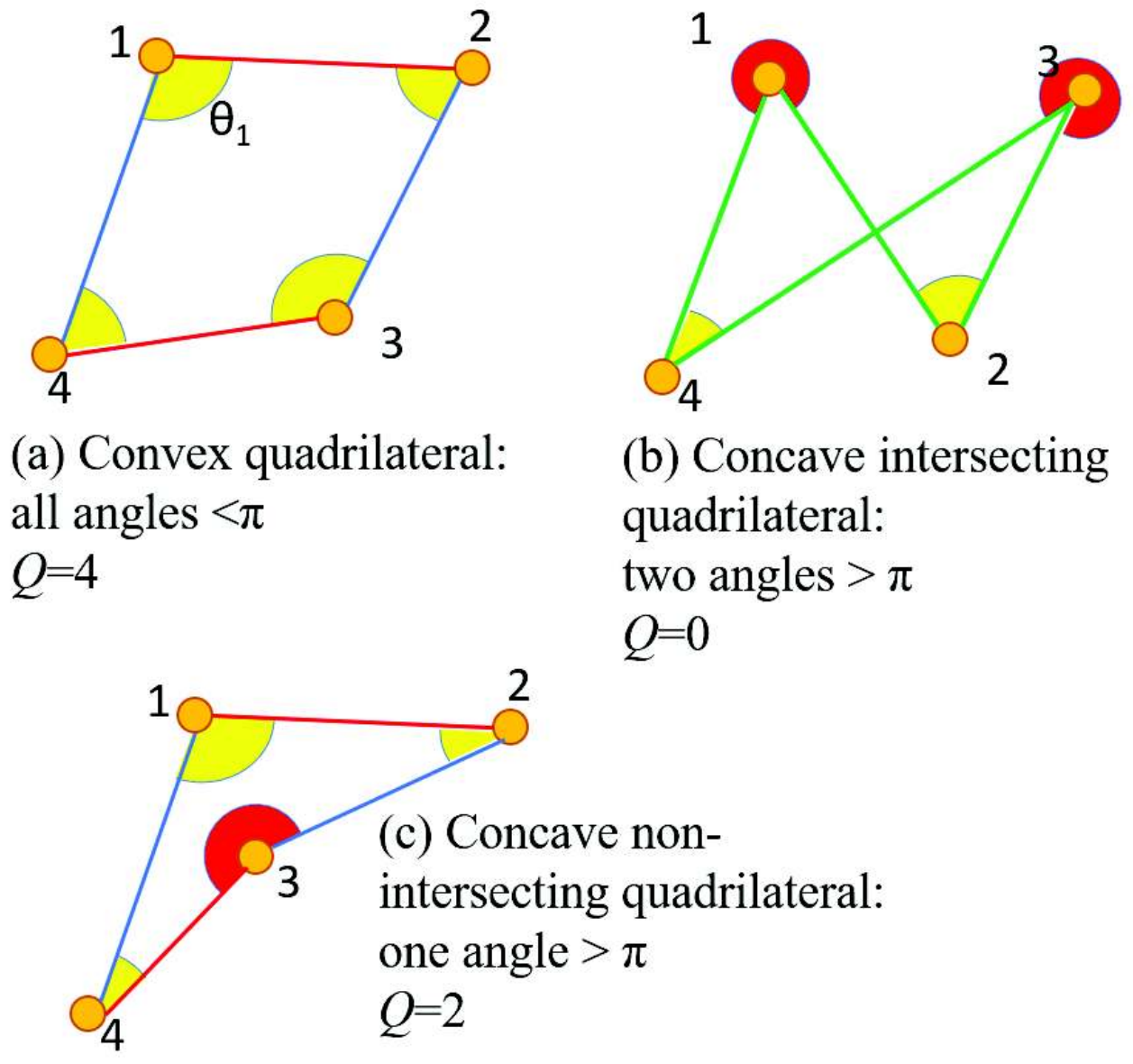

**FIGURE 2.** **All three different occasions of Q value, based on Eq. 5, are depicted. All angles less than $\pi$ are denoted with yellow and the angles greater than $\pi$ are denoted with red.**

For any given quadrilateral the following quantity, $Q$, can be calculated

$$Q = |\sum_{i=1}^{4} sgn(v^i_z)| \quad (5)$$

*Lemma 1:* For any planar quadrilateral, $Q$ obtains one of the following three values: 0, 2, or 4. If Q=4, then the defined quadrilateral is convex, Else if Q=2, then the quadrilateral is concave non-self-intersecting, Otherwise, if Q=0, then the quadrilateral is concave and self-intersecting.

It is easy to verify the Lemma by visualizing the three different types of quadrilaterals, as shown in Fig. 2.

We can utilize the above in random point selection in the proposed $H$-RANSAC, as following. Let the ordered 4-points $p^A$, $p^B$ be randomly selected from image $A$ and $B$, respectively. As a first step at each iteration, the Q-values $Q_A$, $Q_B$ of the two quadrilaterals are calculated using Eq.(5). If $Q_A = Q_B$ then the iteration proceeds, otherwise new points are selected.

### E. DETECTING IMPLAUSIBLE HOMOGRAPHY AT THE END OF EACH ITERATION

During experimentation with our dataset, it was observed that it is possible for an estimated transform $H$ to satisfy a number of inliers and the selected 4 pairs of points to exhibit the same $Q$-value according to the criterion in the previous subsection, however the resulting image transform may still be implausible. Such an example is provided in Fig. 3(bottom), where the selected 4 points in each image constitute convex polygons (with $Q = 4$) as it can be seen in Fig. 3(upper and middle). Six (6) out of the possible 8 inliers were found by the

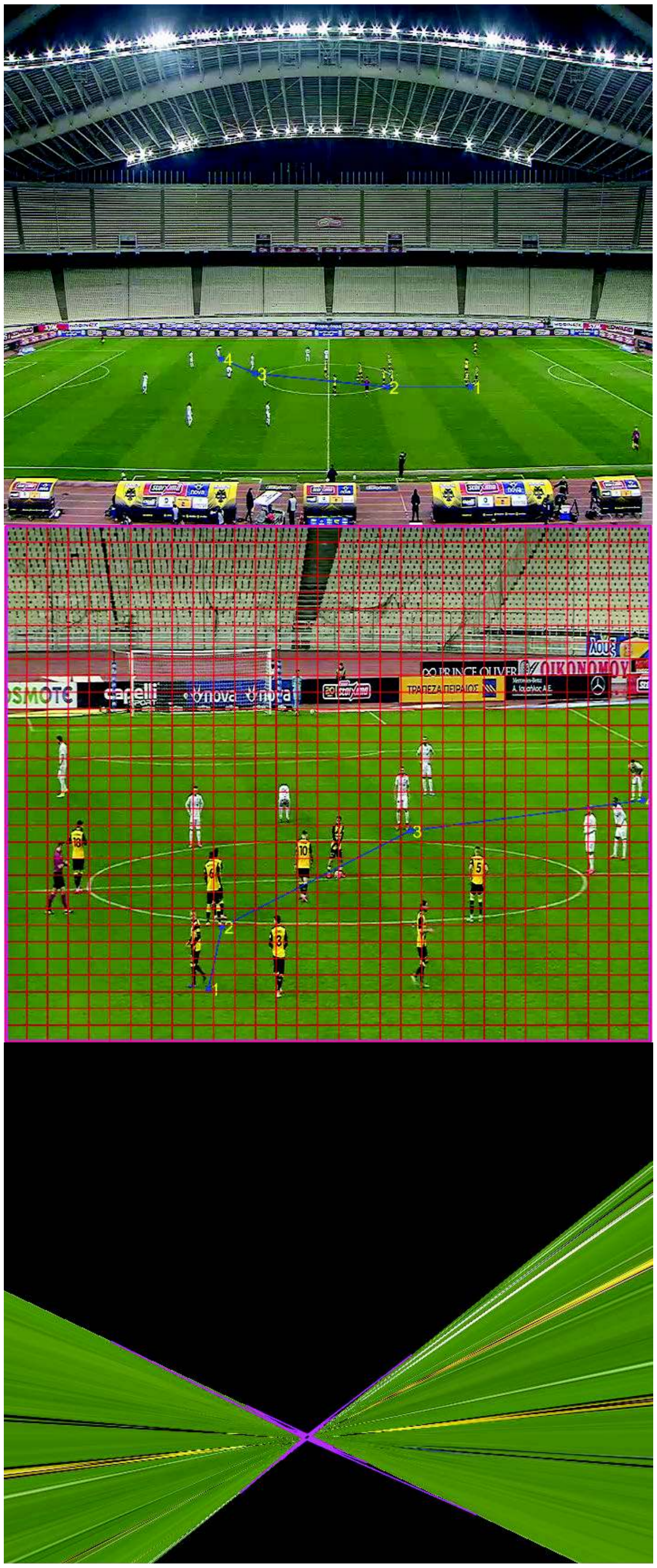

**FIGURE 3.** **Example of wrong homography transformed image (lowest image) detected by the post-hoc criterion $Q_{im} = 0$, despite convex quadrilaterals selected points of both reference frame ($image_A$, shown in the top) and candidate image ($image_B$, shown in the middle), equiv. $Q_A = Q_B = 4$. The transformed grid of the middle image is also superimposed on the lower image to visualize the implausible deformation.**

**Algorithm 2** The Proposed $H$-RANSAC Methodology

**Input:** $P^A$, $P^B$, $n_{iter}$
**Output:** $H$
**Ensure:** transformed image isn't self-intersecting
  $p_1^A \leftarrow$ points of $P^A$ with $c_1$ (class 1) in $image_A$
  $p_2^A \leftarrow$ points of $P^A$ with $c_2$ in $image_A$
  $p_1^B \leftarrow$ points of $P^B$ with $c_1$ in $image_B$
  $p_2^B \leftarrow$ points of $P^B$ with $c_2$ in $image_B$
  $N_1^A \leftarrow$ number of $p_1^A$
  $N_2^A \leftarrow$ number of $p_2^A$
  $N_1^B \leftarrow$ number of $p_1^B$
  $N_2^B \leftarrow$ number of $p_2^B$
  $T \leftarrow \lambda \times \max\{\|P_i^B - P_j^B\|\}, 0 < i, j \leq N^B$
  $n_1 \leftarrow min\{N_1^A, N_1^B\}$
  $n_2 \leftarrow min\{N_2^A, N_2^B\}$
  **if** $n_1 + n_2 \geq 4$ % Eq.(1) **then**
    **while** $\psi_1 < n_{iter}$ **do**
      Determine the classes of the 4 point pairs using Algorithm 1($n_1, n_2$)
      Let $ptsSet_A, ptsSet_B$ be the selected points from images $A$ and $B$
      $\psi_1 \leftarrow \psi_1 + 1$
      calculate $Q_A$ for $ptsSet_A$, according to Eq.5
      calculate $Q_B$ for $ptsSet_B$
      **if** $Q_A \neq Q_B$ **then**
        **continue** (next iteration)
      $H \leftarrow$ estimate $H$ using selected point pairs
      $D_1 \leftarrow$ distance matrix $N_1^A \times N_1^B$ between $p_1^A, Hp_1^B$
      $D_2 \leftarrow$ distance matrix $N_2^A \times N_2^B$ between $p_2^A, Hp_2^B$
      $D \leftarrow \begin{bmatrix} D_1 & inf \\ inf & D_2 \end{bmatrix} \in \mathbb{R}^{N^A \times N^B}$
      Inlier point pairs are determined by thresholding $D$, i.e. $D_{ij} < T$
      $n_{inliers} \leftarrow \#inlier\ pairs$
      **if** $n_{inliers} > 5$ **then**
        $Q_{im} \leftarrow Q\ value\ of\ HE^B$
        **if** $Q_{im} \neq 4$ **then**
          **discard** (current iteration)
        **else**
          $H \leftarrow$ estimate $H$ using inlier point pairs
      **else**
        **discard** (current iteration)

proposed $H$-RANSAC, but the resulting image is distorted Fig. 3(bottom), due to 2 wrong corresponding point pairs. This distortion is evident in the lower image of Fig. 3, where the canonical grid of the middle image is transformed and superimposed. This is a possible occurrence with increased number of points $N^A, N^B$. In order to enable the algorithm to detect such non-plausible homographies, we introduce a post-hoc criterion, after the calculation of $H$ in each iteration and reject implausible homography solutions, as follows.

After the completion of each iteration, the currently estimated $H$ is applied to the four corners of image $B$, $E_i^B, i = 1 \ldots 4$ ordered in a clockwise manner. Since the corner points of the original image form a parallelogram, the transformed corners, $HE_i^B, i = 1 \ldots 4$, considered as a quadrilateral polygon, should be convex under any point of view outside the image corners (i.e. $Q = 4$). Thus, concave transformed quadrilaterals indicate implausible homography.

Every aforementioned step of the proposed method is formally depicted in Algorithm 2.

### F. THE COMPUTATIONAL EFFICIENCY OF H-RANSAC COMPARED TO STATE-OF-THE-ART RANSAC WITH CHEIRAL CONSTRAINTS

Let $\pi_0, \pi_2, \pi_4$ be the probability for selecting 4 points from a single image with $Q = 0, 2, 4$ respectively. Further, let $\hat{\pi}_0, \hat{\pi}_2, \hat{\pi}_4$ be the probabilities of the selected points with corresponding $Q-$ values that satisfy the post-hoc criterion as well. According to our proposed algorithm only a fraction of

$$\phi = \pi_0^2 + \pi_2^2 + \pi_4^2 \tag{6}$$

of the random quadrilateral selection will exhibit equal $Q-$values. Thus the number of 4 point-pairs selections that will proceed in the iteration is expected to be reduced approximately by a factor of $\phi$, compared to a RANSAC without using geometric constraints. The number of iterations of our proposed approach is expected to be a fraction $\hat{\phi}$ of the standard RANSAC:

$$\hat{\phi} = \hat{\pi}_0^2 + \hat{\pi}_2^2 + \hat{\pi}_4^2 \tag{7}$$

A number of works have also utilized the concept of geometric constraints, often referred to as cheiral inequalities. In [44] quadrilateral convexity is expressed in terms of the homography matrix $H$ and the 7 coefficients of the inscribed ellipse inside the minimum rotated quadrilateral bounding the corresponding points. This constraint is not tested in advance for each 4-point pairs, but it is being included in the optimization process, thus the recovered $H$ is based on convex quadrilaterals. This is a different approach that requires a mathematically demanding step at the beginning of the RANSAC. The proposed test in Eq.3 is very similar to the signed area constraint defined in Eq.(9) of [44], or paragraph 3.2.3 *Stage 1b* in [4]. In the latter, quadruples of points are accepted only if 3 out of 4 triangular areas of the quadrilateral have positive sign (equivalently the quadrilateral has $Q = 4$ in our approach). A similar but simplified test is proposed in [45] (subsection III.c), where a 4-point similarity is checked based on the direction of four vectors; this method however is only applicable to very small camera pose changes, such as consecutive video frames.

The aforementioned methods utilize selected points only if $Q = 4$. In our approach, we detect the three different types of quadrilaterals and utilize all of them, provided that the 4 points from each image have the same $Q$ value. Thus, our proposed criterion is less strict and more effective, allowing for a fraction $\phi$ of the selected 4-point pairs to proceed in the current iteration, compared to $\pi_4^2$ for the other methods.

To gain an insight of the expected reduction in the number of iterations and the utilization of the selected point pairs, we performed a numeric simulation by generating random quadrilaterals and calculating the number of occurrences of each $Q-$value and the corresponding probabilities $\pi_0, \pi_2, \pi_4, \phi$. We experimented with different image sizes and 1000s of random quadrilaterals. We also calculated the same quantities for the actual data. The results are shown in Table 1. It can be observed that the probabilities are in very good agreement for the simulation and the actual data. It is evident that, on average the proposed method utilizes 36% of the selected point pairs, compared to 4.8% selected point utilization by the other constraint-based methods. In our proposed method, the final fraction $\hat{\phi}$ of the reduced RANSAC iterations compared to the standard algorithm is very close to $\phi$, since the activations of the post-hoc $Q$-image criterion are very rare (equivalently $\hat{\pi}_i$ is very close to $\pi_i$ for $i = 0, 2, 4$), as can be deduced from Table 1.

**TABLE 1. Arithmetic values of probabilities for the simulation and the actual experimental data.**

| Probability | Simulation | Experimental Results |
|---|---|---|
| $\pi_0$ | 0.4588 | 0.4509 |
| $\hat{\pi}_0$ | 0.4569 | 0.4490 |
| $\pi_2$ | 0.3095 | 0.3370 |
| $\hat{\pi}_2$ | 0.3018 | 0.3366 |
| $\pi_4$ | 0.2443 | 0.2233 |
| $\hat{\pi}_4$ | 0.2413 | 0.2230 |
| $\pi_4^2$ | 0.057 | 0.048 |
| $\phi$ | 0.3627 | 0.3556 |

The theoretically expected number of iterations as a function of points per image will be derived in the next subsection.

### G. ESTIMATION OF THE NUMBER OF ITERATIONS FOR THE PROPOSED H-RANSAC

First we will estimate the expected number of iterations, $n_{iter}$ of the proposed generalized RANSAC for two sets of points ($N^A$ points in $P^A$ and $N^B$ points in $P^B$) of a single class, without the application of geometry-based tests. Let $P$ be the subset of points in $P^A$ that correspond to points in $P^B$ and vice versa and $k = \#P$ be the number of correct point correspondences between $P^A$ and $P^B$. Further, let us denote by the $C_b^a = \frac{a!}{(a-b)!b!}$ number of combinations of $b$ from a total of $a$ points.

The probability of selecting 4 points from $P^A$ that belong to $P$ is $p_1 = \frac{C_4^k}{C_4^{N^A}}$. Similarly, the probability of selecting 4 points from $P^B$ that belong to $P$ is $p_2 = \frac{C_4^k}{C_4^{N^B}}$. The probability to select the same 4 common points from $P^A$ and $P^B$ with the same order is given by $p_0 = \frac{p_1 p_2}{C_4^k 4!}$. If we combine the result from the subsection II-D on the above equation, then $p_0 = \frac{p_1 p_2}{0.36 \times C_4^k 4!}$. Finally, if $p_r$ is the preset probability of correct 4-point pairs selection, then the required number of iterations is given by

$$n_{iter} = \left\lceil \frac{log_{10}(1 - p_r)}{log_{10}(1 - p_0)} \right\rceil, \tag{8}$$

**TABLE 2. The expected number of iterations $n_{iter}$ of the proposed *H*-RANSAC for different number of points per image, with $p_r = 0.95$.**

| $N_1^A$ | $N_2^A$ | $k_1$ | $N_1^B$ | $N_2^B$ | $k_2$ | $n_{iter}$ |
|---|---|---|---|---|---|---|
| 6 | 0 | 4 | 6 | 0 | 0 | 5822 |
| 3 | 3 | 2 | 3 | 3 | 2 | 348 |
| 8 | 0 | 4 | 8 | 0 | 0 | 126826 |
| 4 | 4 | 2 | 4 | 4 | 2 | 5589 |
| 10 | 0 | 4 | 10 | 0 | 0 | 1141144 |
| 5 | 5 | 2 | 5 | 5 | 2 | 43137 |

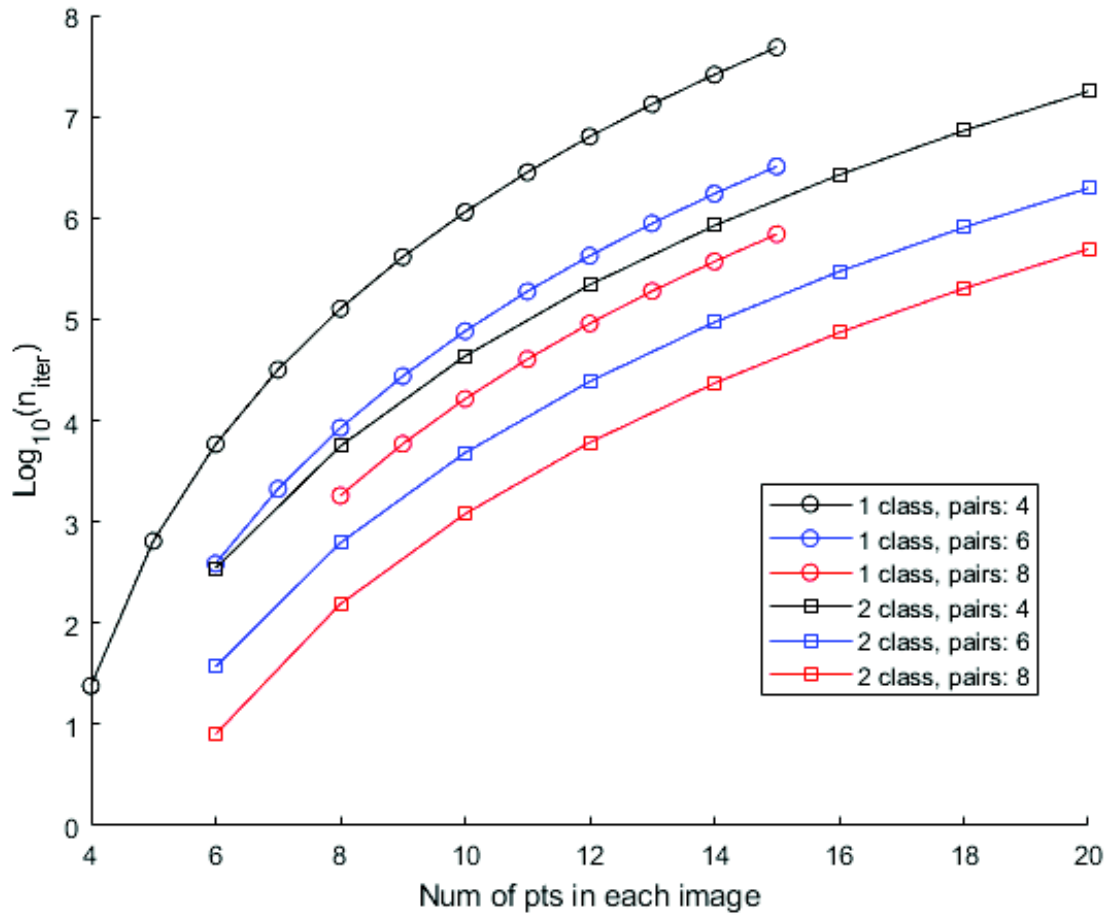

**FIGURE 4. Logarithm of theoretically expected iterations ($n_{iter}$) for different $N_1^A, N_2^A, N_1^B, N_2^B$.**

where [.] denotes the rounding to the nearest integer. Now we elaborate the above calculation for points that belong to two classes. As already defined in subsection II-B, $N_1^A, N_2^A$ are the number of points for $class_1, class_2$ in image A and $N_1^B, N_2^B$ the number of points for $class_1, class_2$ in image B. Further, $N^A = N_1^A + N_2^A$ and $N^B = N_1^B + N_2^B$ are the total number of points in $P^A$ and $P^B$, respectively.

Let us also denote by $P_1$ the subset of points of $class_1$ in $P^A$ that correspond to points in $P^B$, and $P_2$ the subset of points of $class_2$ in $P^A$ that correspond to points in $P^B$ and $k_1 = |P_1|$ and $k_2 = |P_2|$. Each time 4 points are selected from each image for the calculation of homography, $e_1$ of which will belong to $class_1$ and $e_2 = 4 - e_1$ will belong to $class_2$, as described above. The probability of selecting 4 points from $P^A$ that belong to $P_1$ is given by $p_1 = \frac{C_{e_1}^{k_1}}{C_{e_1}^{N_1^A}} \frac{C_{4-e_1}^{k_2}}{C_{4-e_1}^{N_2^A}}$. Similarly, the probability of selecting 4 points from $P^B$ that belong to $P_2$ is $p_2 = \frac{C_{e_1}^{k_1}}{C_{e_1}^{N_1^B}} \frac{C_{4-e_1}^{k_2}}{C_{4-e_1}^{N_2^B}}$. Finally, the probability to select the same 4 common points from $P^A$ and $P^B$ with the same order is $p_0 = \frac{p_1 p_2}{C_{e_1}^{k_1} C_{4-e_1}^{k_2} e_1! (4-e_1)!}$. If we combine the above result with the point selection refinement (subsection II-D), then the probability $p_0$ is further increased:

$$p_0 = \frac{p_1 p_2}{\phi \times C_{e_1}^{k_1} C_{4-e_1}^{k_2} e_1! (4 - e_1)!} \tag{9}$$

The required number of iterations $n_{iter}$ is again given by Eq.(8).

Table 2 provides the number of iterations *niter* for $p_r = 0.95$ for 6, 8 and 10 points per image, with 4 point pairs (thus fraction of outliers equal to 0.33, 0.5 and 0.66 respectively). Results are presented for the case of a single class ($N_2^A = N_2^B = 0$), as well as for two classes of points ($N_2^A = N_1^A, N_2^B = N_1^B$). the value of $n_{iter}$ ranges from $10^3$ to $10^5$ for 2-class points, achieving a 20-fold reduction compared to single class points. More detailed calculations are provided in the graph of Fig. 4.

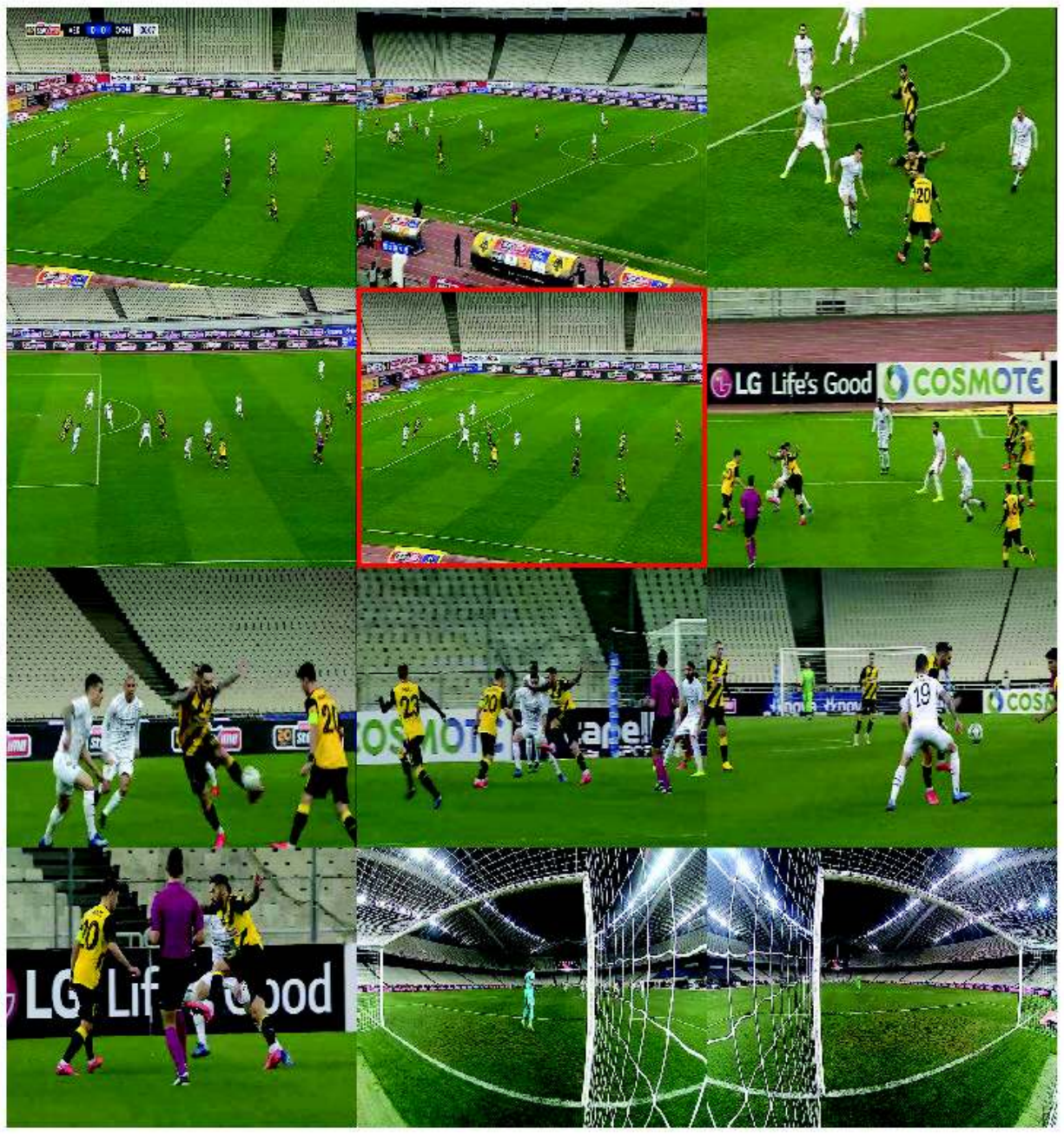

**FIGURE 5. Random example from our dataset of 12 simultaneous frames. The reference frame is annotated in red.**

## III. RESULTS

### A. DESCRIPTION OF THE DATASET

The proposed methodology was tested on football game images from an extensive dataset, created and annotated by our team that has not been made publicly available yet. The dataset contains 8446 images of size 720 × 576, that were captured during the duration of a number of football games. Except for the reference camera frame, eleven (11) more frames were captured for each moment (timestamp) of the games, from an equal number of cameras with different positions and radically different orientations, creating 11 possible image pairs that can be used for homography estimation. The number of eligible pairs of frames, according to the criterion of Eq.1 (existence of enough persons in the playing field) is equal to 2312 with 528 unique reference frames. Equiv-alently, on average 4.4 frames of the same time stamped can be matched with the corresponding reference frame. The pre-trained YOLOv5 [47] was used to create a file that contains the position and class (team) of the players for each image. The files were reviewed by human annotators that identify each football player. Fig. 5 depicts a typical example of the available images for a single timestamp. The reference frame that captures the main part of the field is also highlighted.

**TABLE 3.** **The effect of $\lambda$ parameter on the performance of $H$-RANSAC.**

| | Wrong point pairs | $\lambda$=0.005 | $\lambda$=0.01 | $\lambda$=0.02 |
|---|---|---|---|---|
| Number of Frame Pairs | 0 | 713 | 939 | 1063 |
| | $\geq 1$ | 218 | 392 | 572 |
| Frame pairs processed | | 931 | 1331 | 1635 |
| Q criterion activation | | 32 | 250 | 1337 |

**TABLE 4.** **Average number of point pairs per image pair.**

| | $\lambda$=0.005 | $\lambda$=0.01 | $\lambda$=0.02 |
|---|---|---|---|
| Ground truth | 14.3 | 13.4 | 12.6 |
| Correctly found | 6.6 | 8.03 | 9.6 |
| Wrong | 0.59 | 0.85 | 1.13 |

### B. QUANTITATIVE RESULTS

The proposed $H$-RANSAC is executed for each one of the 2312 frame pairs (with the reference frame at that timestamp being the reference image).

Since the points of interest (players on the football field) have already been automatically annotated in the frames using the Yolo-based player detector (as will be discussed in III-A), the correct point pairs (i.e. *inliers*) can be checked for each method by applying the estimated matrix $H$ to the annotated (ground truth) points, calculating their reprojection error and applying the same threshold as in Eq.(3).

Table 3 provides details of the execution for three different values of the $\lambda$ parameter (lower $\lambda$ corresponds to lower threshold $T$ for determining inlier point pairs). More specifically, the average number of frame pairs with at least four (4) identified point pairs, of which a) 0 wrongly identified point pairs and b) one or more wrongly identified points pairs are reported. The homography transform between any frame pair with least 4 identified point pairs, of 0 wrong point pairs, is bound to be correct according to theory. In any other case (i.e., less than 4 identified point pairs, or 1 or more wrongly identified point pairs) the frame pair is considered as wrongly aligned. In addition, the number of processed frame pairs and the average number of activation of the $Q$ criterion per frame pair is also given, which increases as the value of the threshold $T$ relaxes. This indicates the importance of the proposed post-hoc criterion.

Table 4 provides further information about the efficiency of the proposed algorithm. More specifically, the average number of ground truth point pairs per image pair and the average point pairs correctly and wrongly discovered per image pair for three different values of $\lambda$.

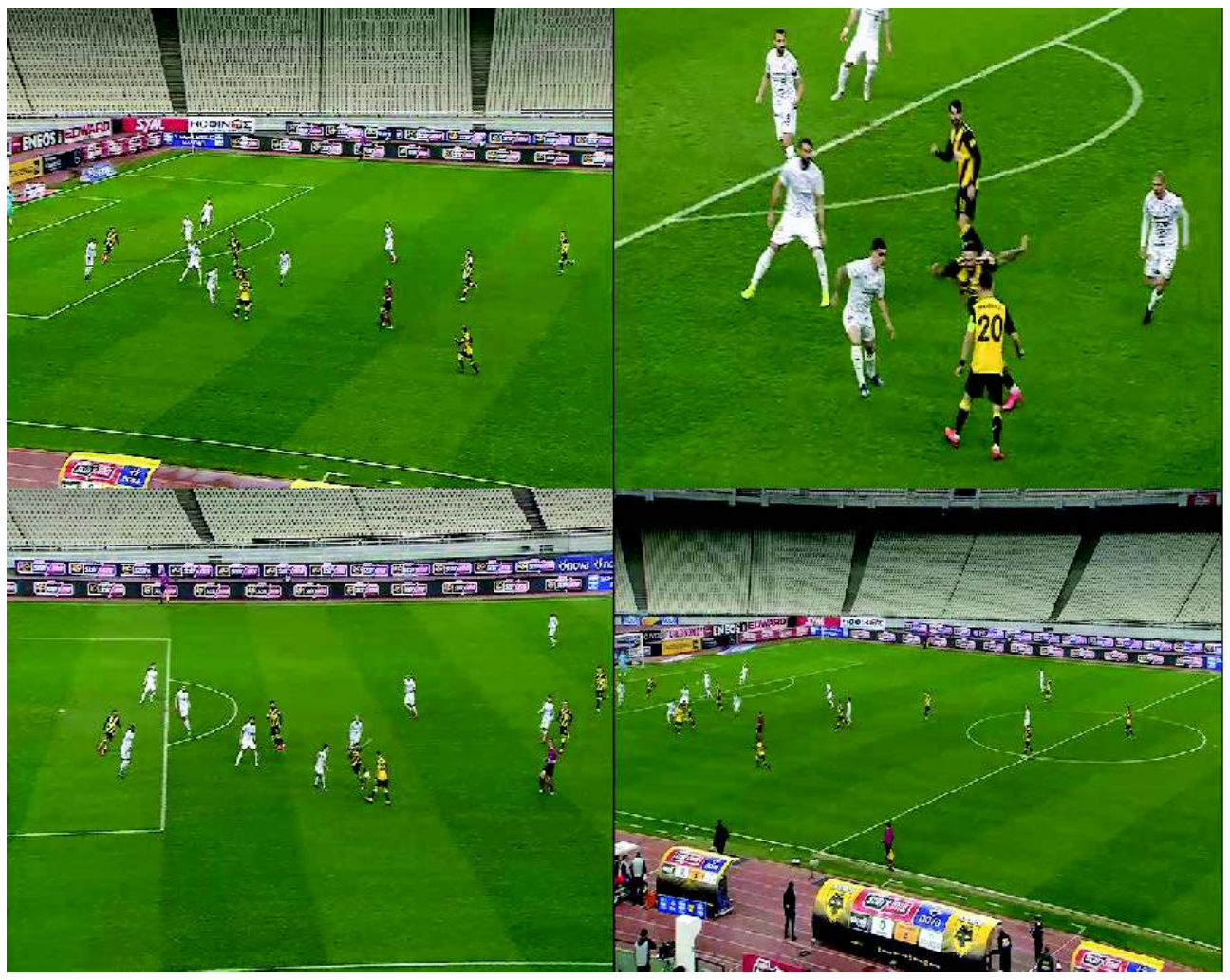

**FIGURE 6.** **Example of given frames for the particular (00_41_26) timestamp.**

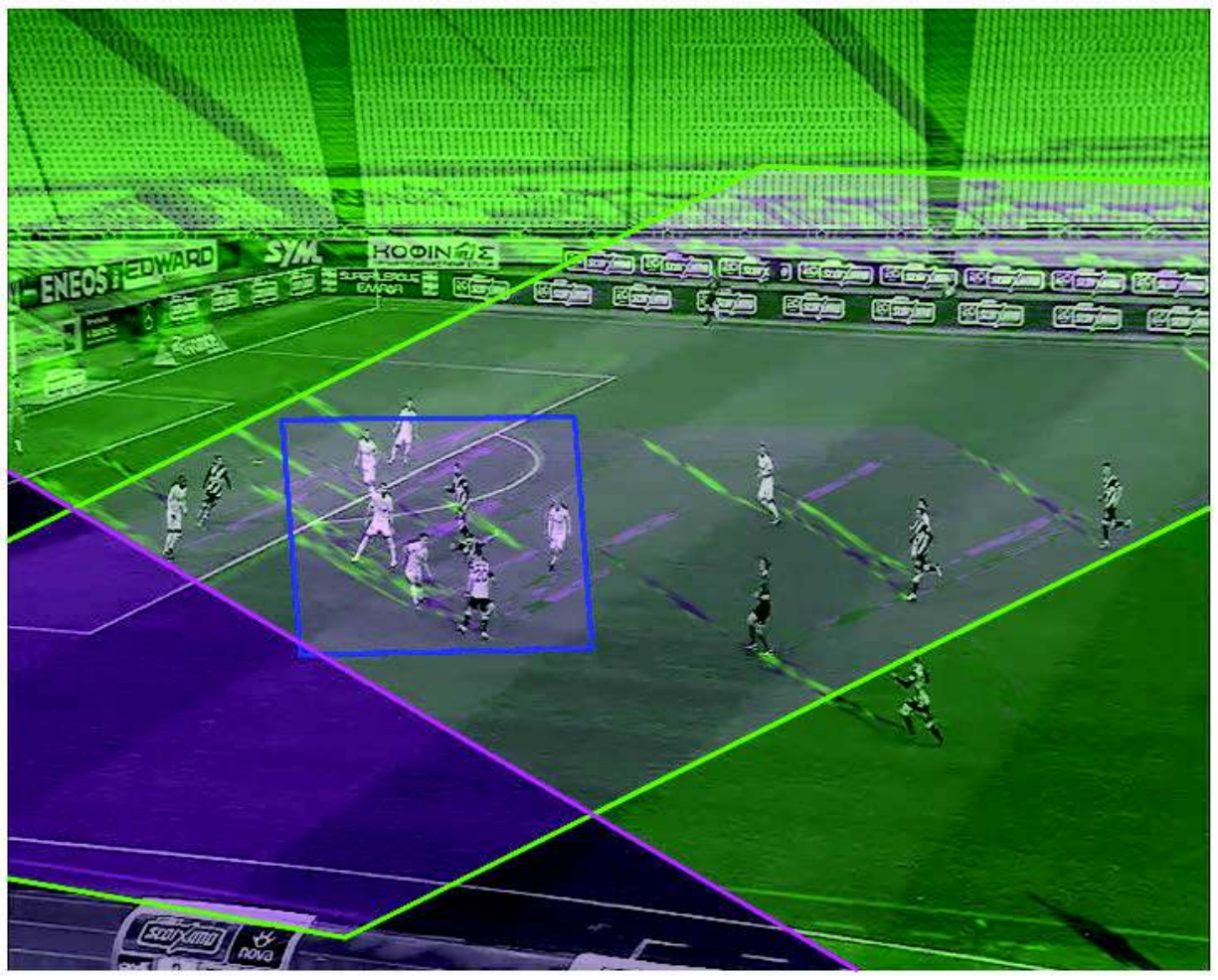

**FIGURE 7.** **Three overlaid transformed images on reference frame for 00_41_26 timestamp.**

Averaging is performed with respect to the number of pairs of frames the proposed algorithm managed to recover a homography matrix.

As expected, increasing the values of $\lambda$ causes a slight increase of the number of frame pairs with 0 wrongly identified points pairs and a steeper increase of the number of frames with one or more wrongly identified point pairs. The average number of point pairs discovered per frame pair increases from 6.6 (with 0.59 wrong pairs) to 9.6 (with 1.13 wrong point pairs). These results are easily explained, considering that increasing $\lambda$ makes point pair selection less strict. Consequently, the number of activations (per frame pair) of the post-hoc $Q$-value test increases from just 32 to 1337.

On average, using $\lambda = 0.02$ (a more relaxed criterion for defining inliers) results in almost 10 detected inliers per image pair, with one of them being wrong. Using stricter values of $\lambda$ decreases the number of detected inliers to approximately 8 and 6, with an average of 0.7 wrong inliers. As expected, the few wrong inliers occur in cases where the points (players) are distributed closely to each other.

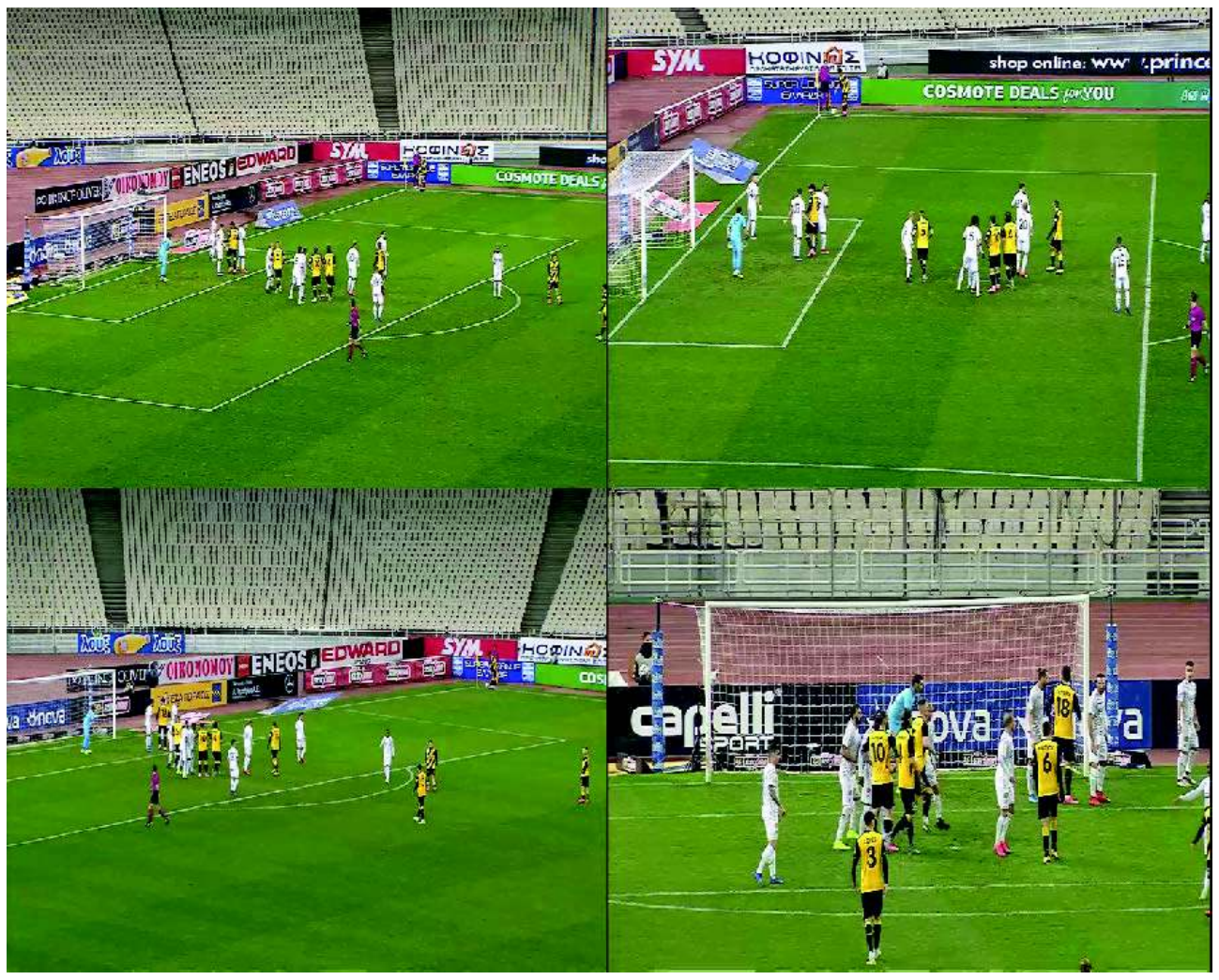

**FIGURE 8.** Example of given frames for the particular (01_07_42) timestamp.

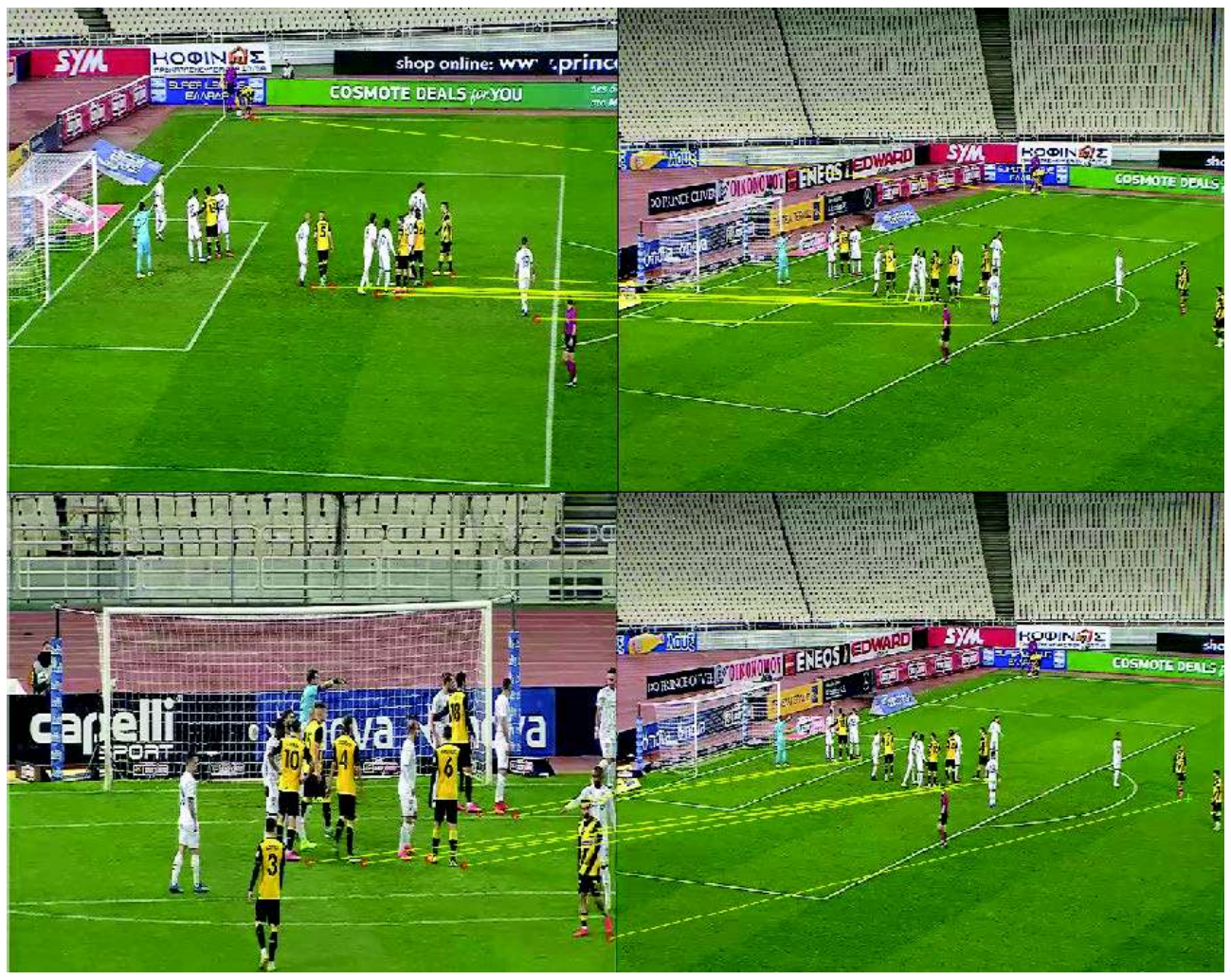

**FIGURE 10.** The detected point (players) correspondences for the frames (magenda) of Fig.8.

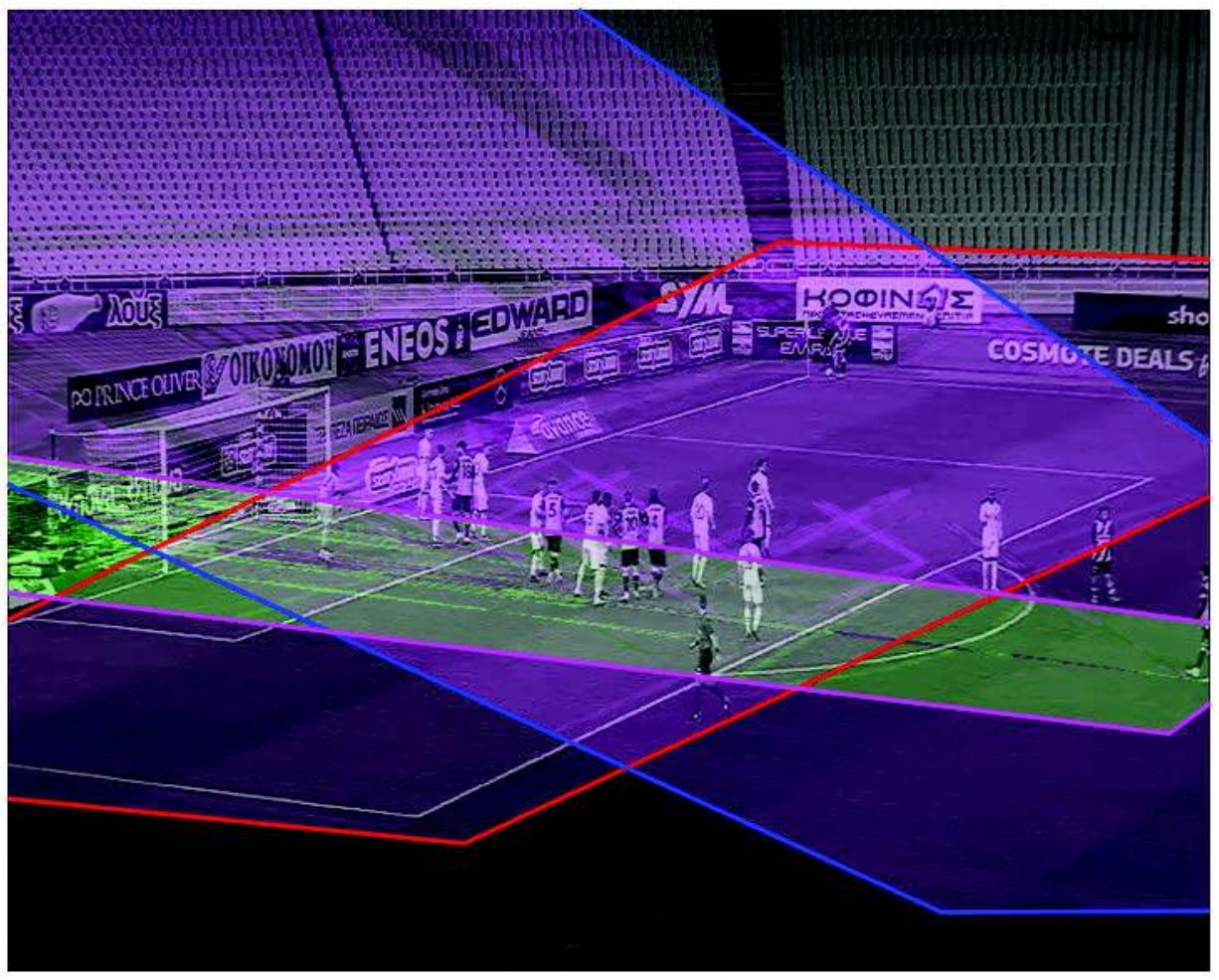

**FIGURE 9.** The reference frame and 3 overlaid matched images for 01_07_42 timestamp (Fig. 8), correctly transformed, despite the extreme camera angle and zoom factor difference.

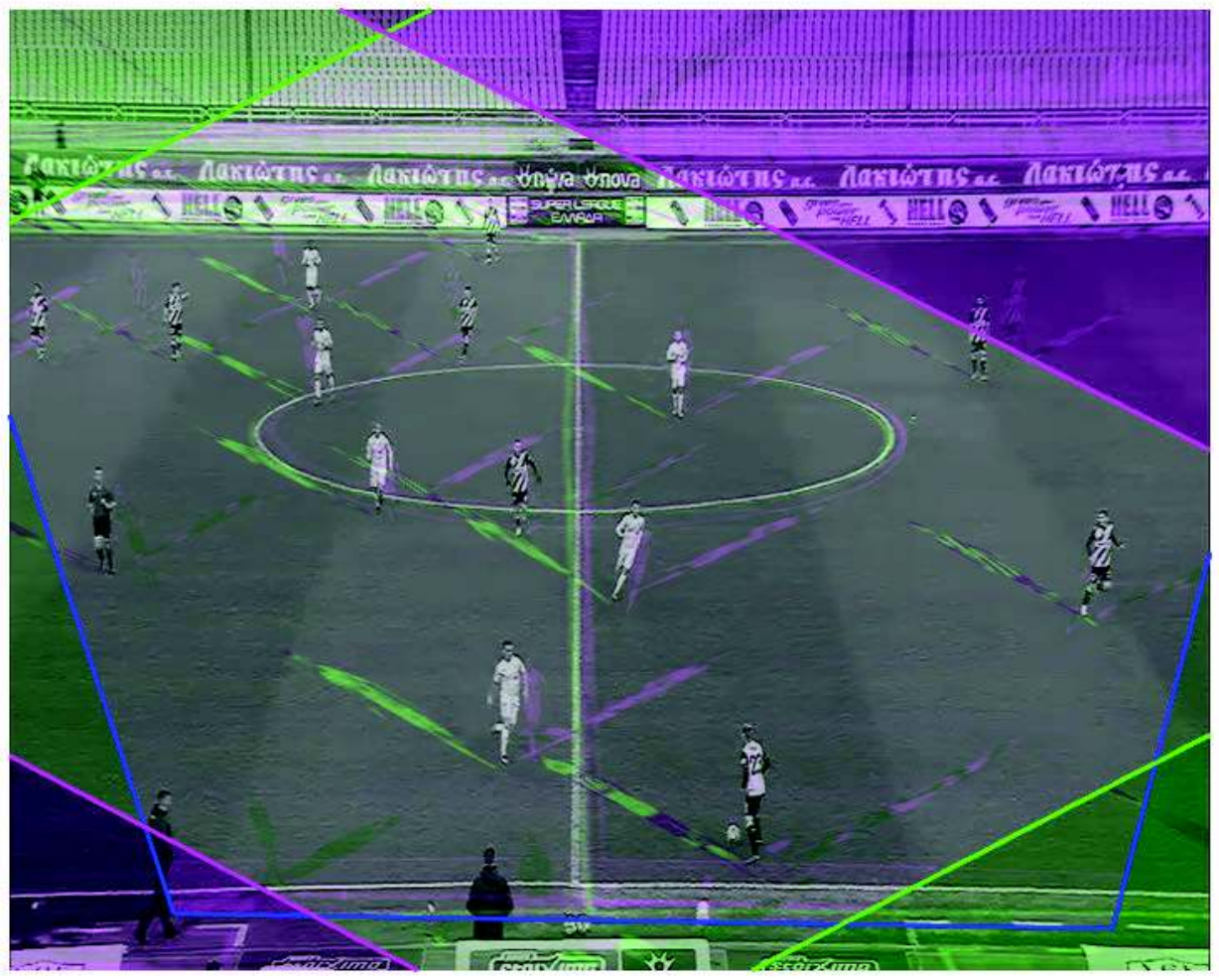

**FIGURE 11.** Typical example of a reference frame with 3 overlaid transformed images.

Fig. 6 depicts the reference frame (upper left) and three other frames with the same timestamp. The homography-transformed frames are overlaid on the reference frame in Fig. 7. Similar results are shown in Figures 8 and 9 for another timestamp. Fig. 10 depicts the recovered point correspondences between the reference frame and the most oblique transformed frames in Fig. 9. The radical changes in orientation and zoom factor can be appreciated. Please note that since homography is recovered using the plane of the field, only objects lying on the field are aligned with the reference frame (e.g. the feet of the players). Figure 11 provides another typical example of transformed frames overlaid on the reference camera frame, using the estimated homography matrix $H$.

Let us define the acceleration of the proposed algorithm, achieved by testing the $Q$ value before each iteration, as the ratio of all iterations over the actually tested iterations (more specifically $\frac{\psi_1}{\psi_2}$ where $\psi_1$, $\psi_2$ are the counters in the pseudocode of Algorithm 2). Figure 12 shows the boxplot of the acceleration induced by the cheiral constraints for all successfully matched image pairs. The mean value of the achieved acceleration is approximately 2.76, which is in agreement with the estimate in subsection II-E.

### C. COMPARISON WITH STATE-OF-THE-ART METHODS

Several approaches that combine local feature extraction and point matching before applying a RANSAC variant, are also being evaluated. We employed the well known SIFT and ORB methods, as well as the deep learning SuperPoint [12], SOSNet [30] and SGA-net [29] methods to detect salient points, extract their feature vector and decide candidate image pairs. DKM [43] and GeoFormer [39] also return pairs of matching points, without feature vectors. The aforementioned methods were combined with well-known RANSAC variants: MAGSAC++ [6], VSAC [7], Graph-Cut [8] and DEGENSAC [10]. The method of bilateral functions [19] has also been included.

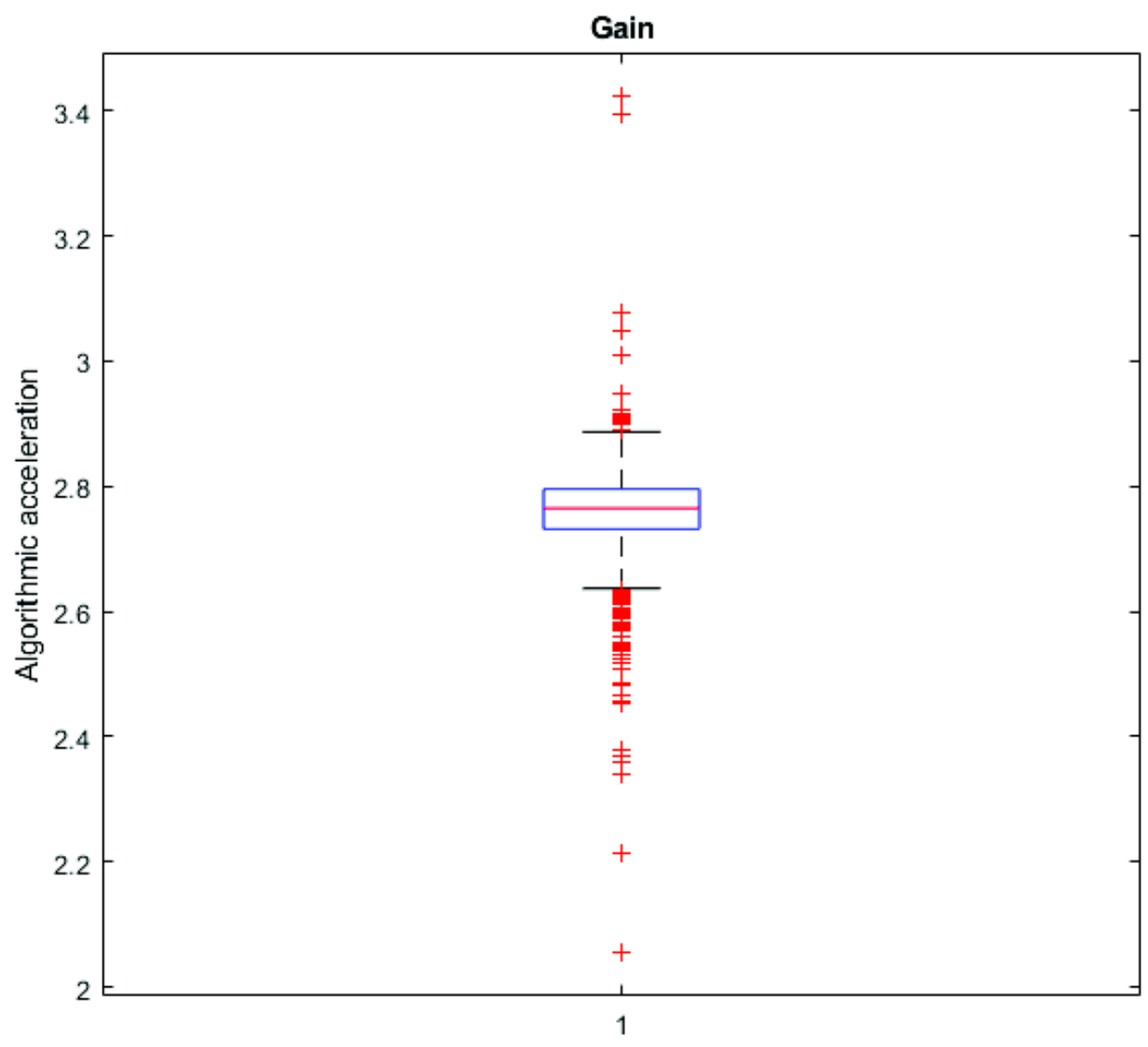

**FIGURE 12.** Boxplot of the acceleration achieved by the cheiral constraints of $H$-RANSAC, over the image pairs.

**TABLE 5.** The number of frame pairs correctly and wrongly aligned, for all methods under comparison.

| Methodology | Frame Pairs Correctly aligned | Frame Pairs Wrongly aligned | Total |
|---|---|---|---|
| SIFT [16]+MAGSAC++ [6] | 41 | 361 | 402 |
| ORB [18]+MAGSAC++ [6] | 40 | 280 | 320 |
| SuperPoint [12]+MAGSAC++ [6] | 131 | 818 | 949 |
| DKM [43]+MAGSAC++ [6] | 125 | 1069 | 1194 |
| GeoFormer [39]+MAGSAC++ [6] | 139 | 1041 | 1180 |
| SOSNet [30]+MAGSAC++ [6] | 94 | 353 | 447 |
| SGA-Net [29]+MAGSAC++ [6] | 97 | 442 | 537 |
| SIFT [16]+VSAC [7] | 42 | 368 | 410 |
| ORB [18]+VSAC [7] | 37 | 271 | 308 |
| SuperPoint [12]+VSAC [7] | 125 | 797 | 922 |
| GeoFormer [39]+VSAC [7] | 133 | 1045 | 1178 |
| SOSNet [30]+VSAC [7] | 88 | 366 | 454 |
| SGA-Net [29]+VSAC [7] | 99 | 446 | 545 |
| SIFT [16]+Graph-Cut [8] | 43 | 354 | 397 |
| ORB [18]+Graph-Cut [8] | 37 | 271 | 308 |
| SuperPoint [12]+Graph-Cut [8] | 127 | 794 | 921 |
| GeoFormer [39]+Graph-Cut [8] | 138 | 1027 | 1165 |
| SOSNet [30]+Graph-Cut [8] | 91 | 357 | 448 |
| SGA-Net [29]+Graph-Cut [8] | 98 | 439 | 537 |
| SIFT [16]+DEGENSAC [10] | 42 | 548 | 590 |
| ORB [18]+DEGENSAC [10] | 40 | 339 | 379 |
| Bilateral Functions [19] | 0 | 278 | 278 |
| Proposed $H$-RANSAC, $\lambda = 0.01$ | 939 | 392 | 1331 |

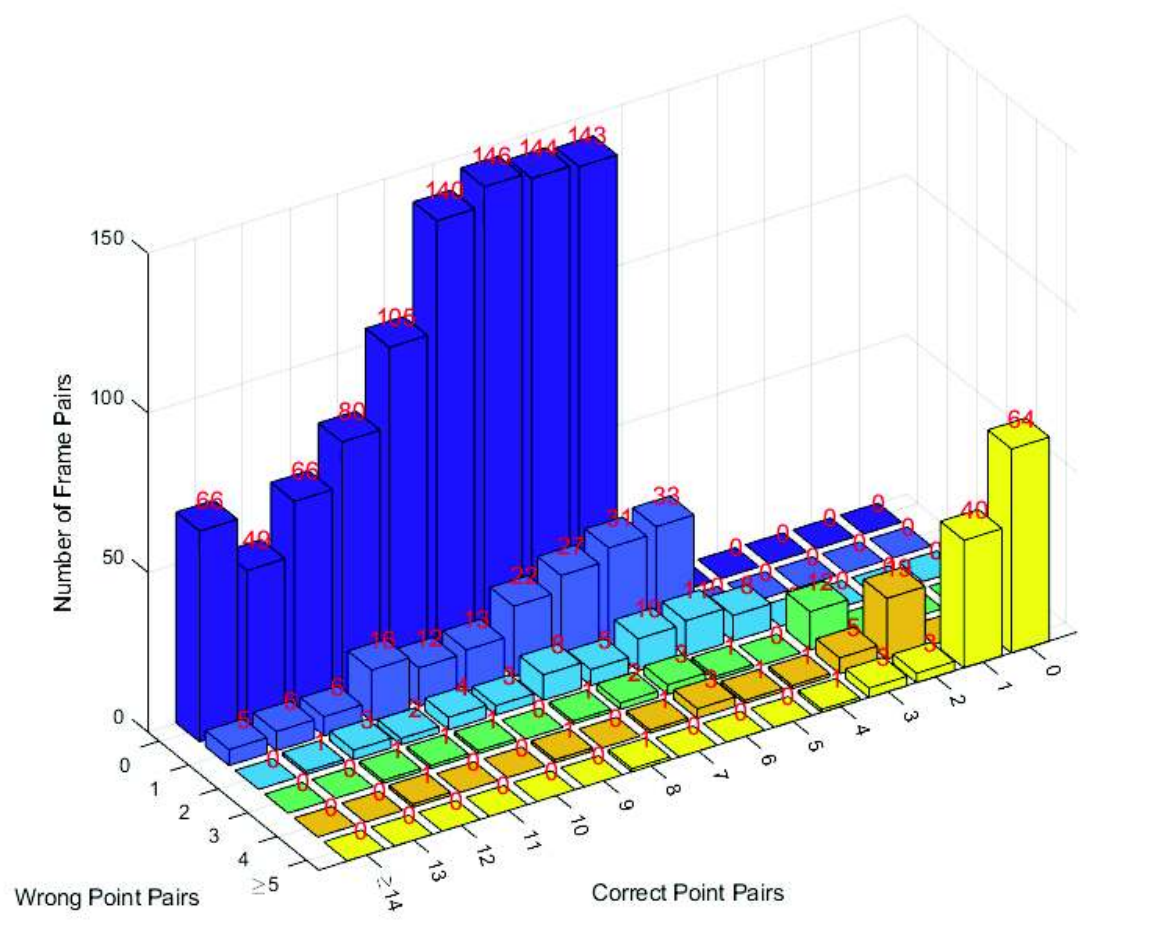

(a)

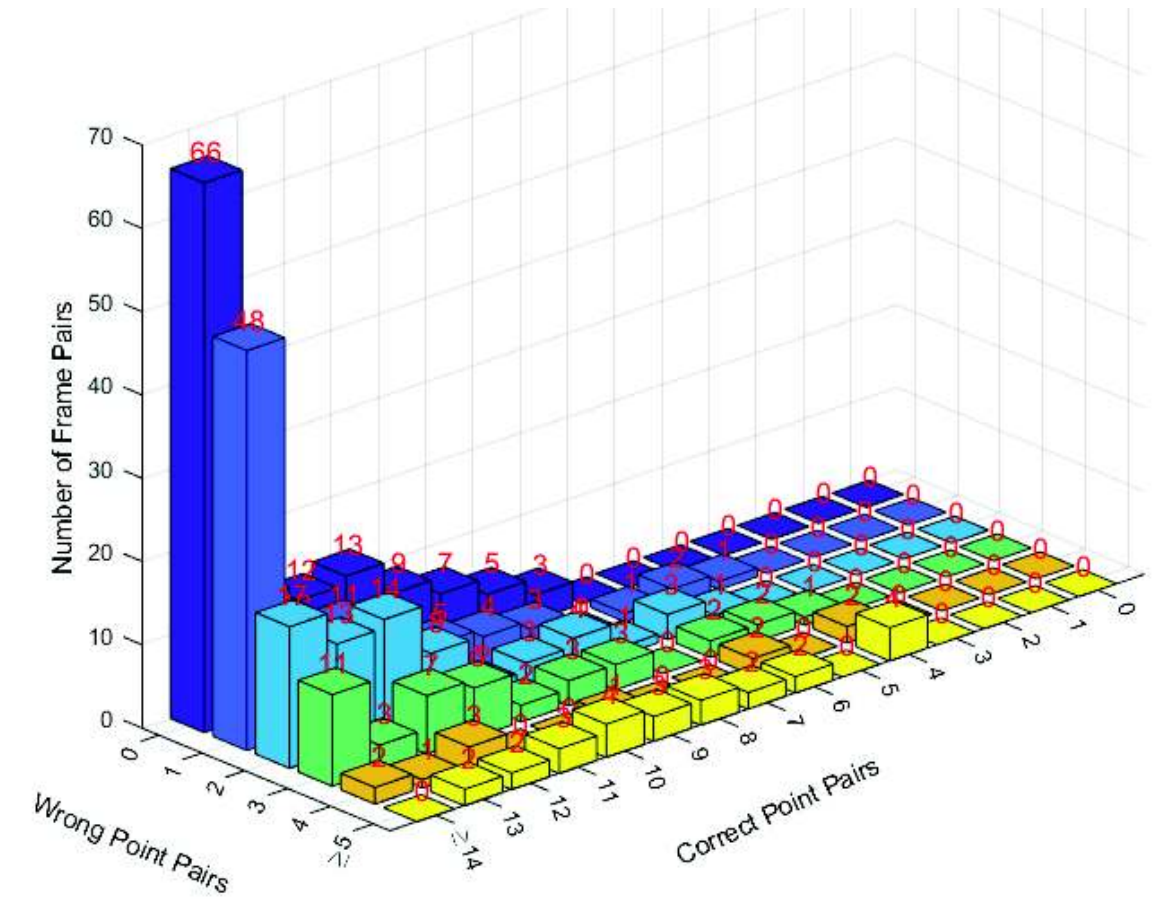

(b)

**FIGURE 13.** (a)The number of image pairs considering the number of correct and wrong point pairs for the proposed $H$-RANSAC for $\lambda$=0.01.(b)The number of image pairs considering the number of correct and wrong point pairs, for Graph-Cut [8] using SuperPoint [12] as descriptor.

Table 5 compares the methods in terms of the number of correctly and wrongly aligned frame pairs. A frame pair is considered correctly aligned if at least 4 of the ground truth points have been correctly paired and none has been incorrectly paired. Finally the rightmost column shows the number of frame pairs that each method managed to process and obtain $H$ matrix (correctly or not).

As it can be observed in Table 5 the proposed $H$-RANSAC achieves the highest number of correctly aligned image pairs (939 out of 1331 frame pairs), outperforming by far the best of the remaining methods (i.e. Superpoint+MAGSAC with 131 out of 939 frame pairs, GeoFormer+MAGSAC with 139 and DKM+MAGSAC with 125 correct image pairs). $H$-RANSAC exhibits low number of wrongly matched image pairs: 392 out of 1331 processed image pairs, whereas the aforementioned methods have a much higher number

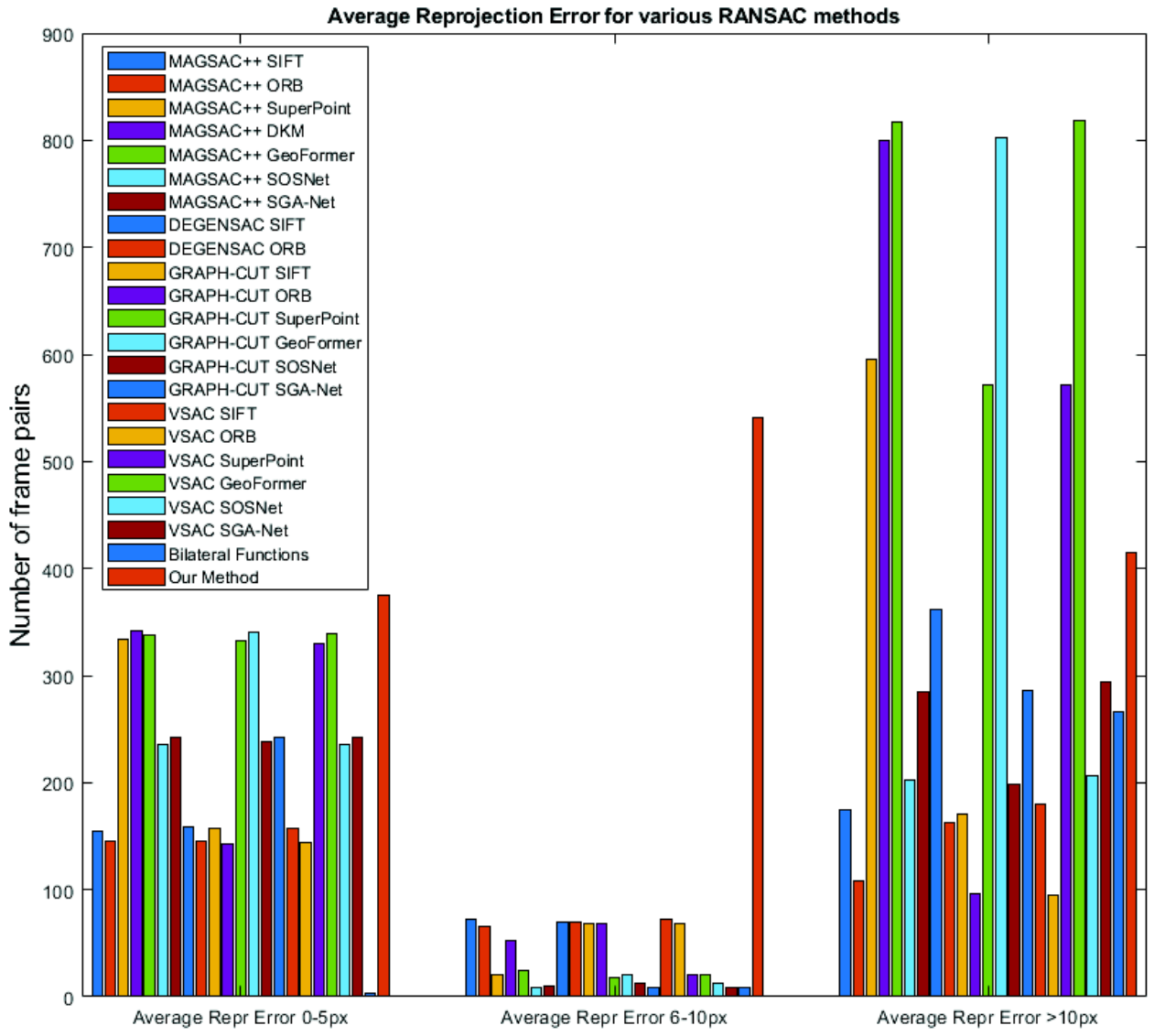

**FIGURE 14.** **The number of frame pairs with average reprojection error in range [0,5], [6,10] and > 10 (pixels) by each method (left center and right bar plot respectively).**

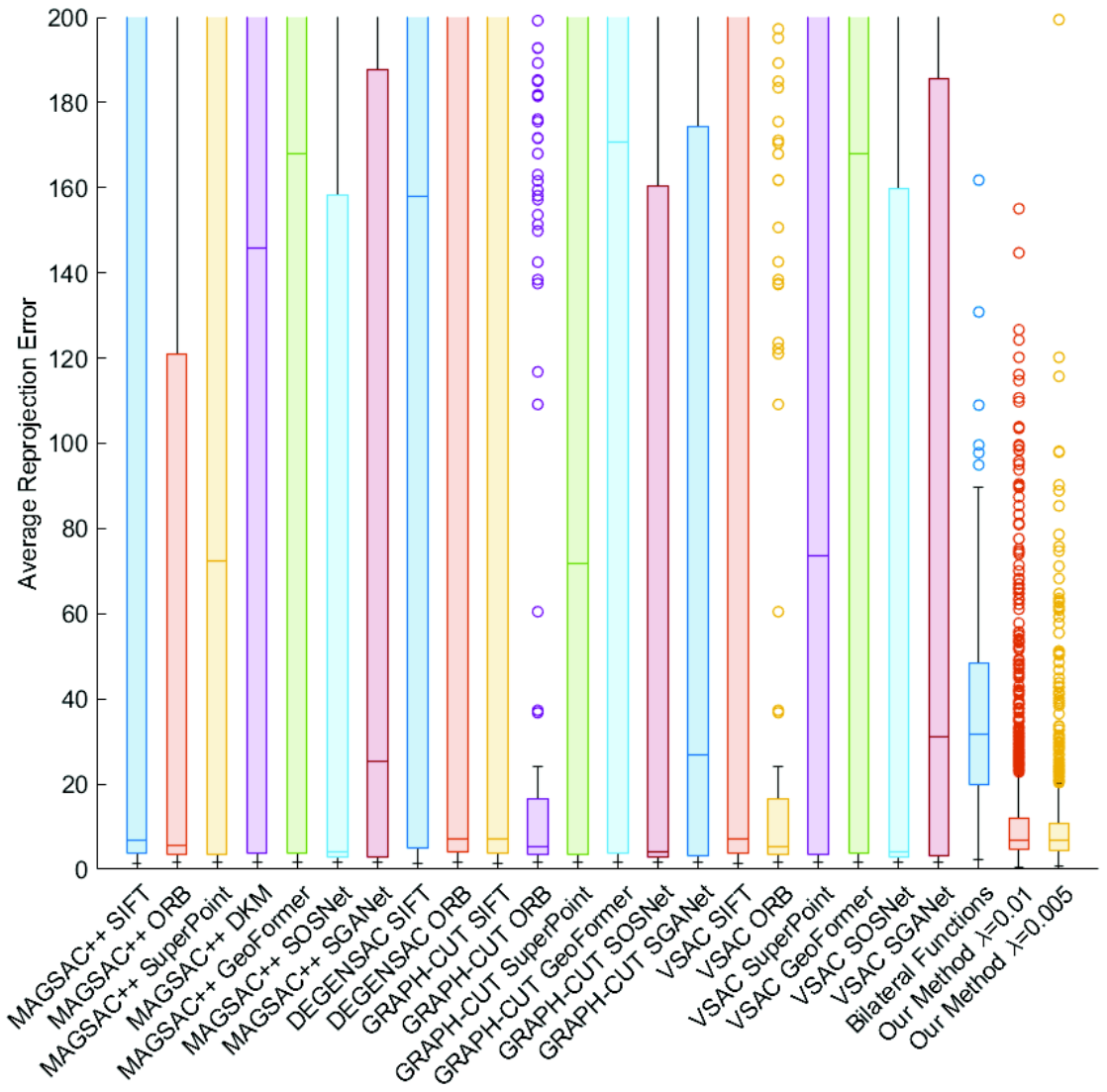

**FIGURE 15.** **Reprojection error of correct point pairs, achieved by the proposed RANSAC for $\lambda = 0.01$ and $\lambda = 0.005$, compared to the state-of-the-art methods. Please note that the number of frames processed by each method differ radically.**

(MAGSAC with Superpoint+ 818 out of 949 processed image pairs, GeoFormer 1041 out of 1180 and DKM with 1069 out of 1194 frame pairs). It should also be mentioned that deep learning-based point detection and feature extraction methods (Superpoint, SOSNet and SGA-Net) outperform the classic ones (SIFT and ORB). The method of ORB+MAGSAC achieved a very low number of wrongly aligned image pairs (280), however this method managed to recover $H$ for only 320 pairs of frames.

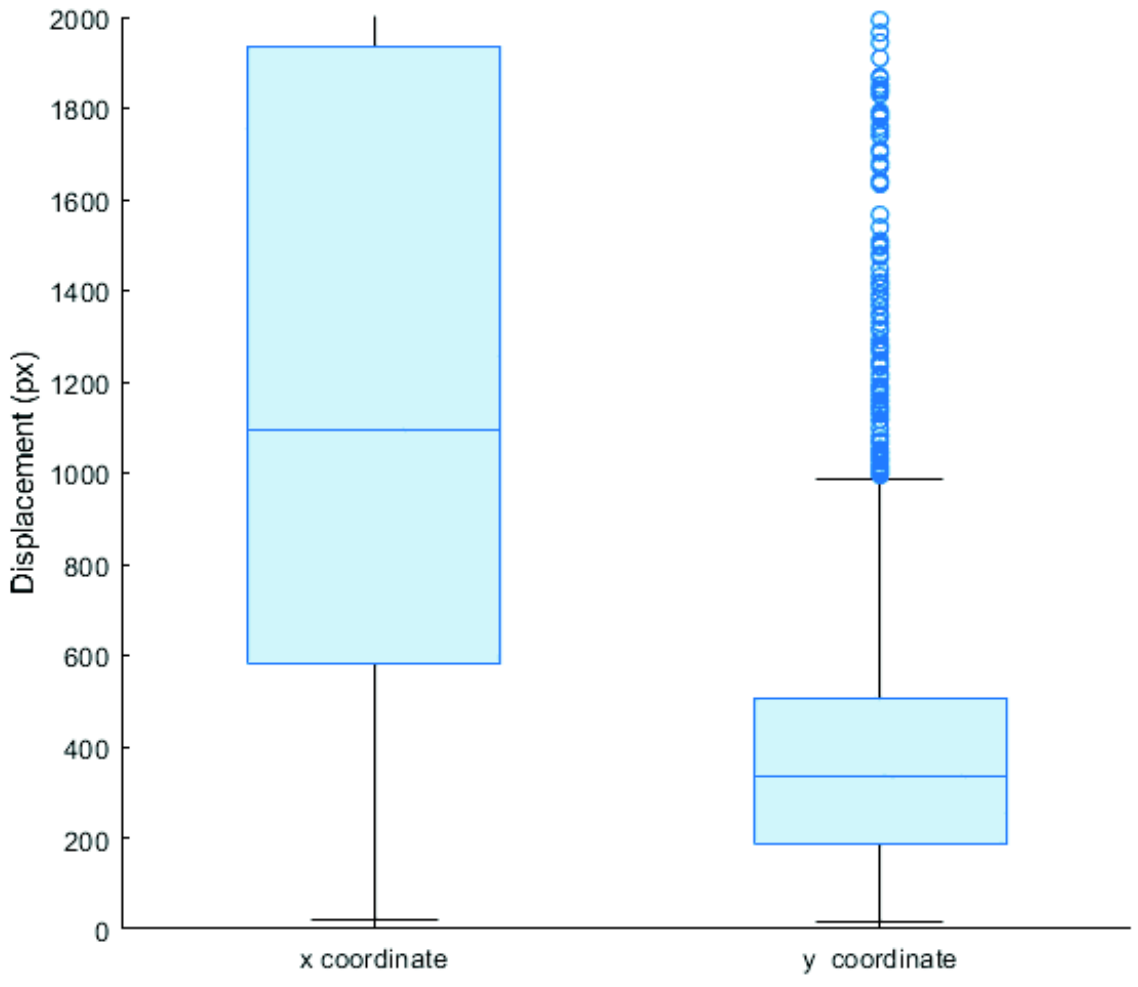

**FIGURE 16.** **Boxplots of the equivalent image corner displacements, according to the homography transform of each image pair in the dataset.**

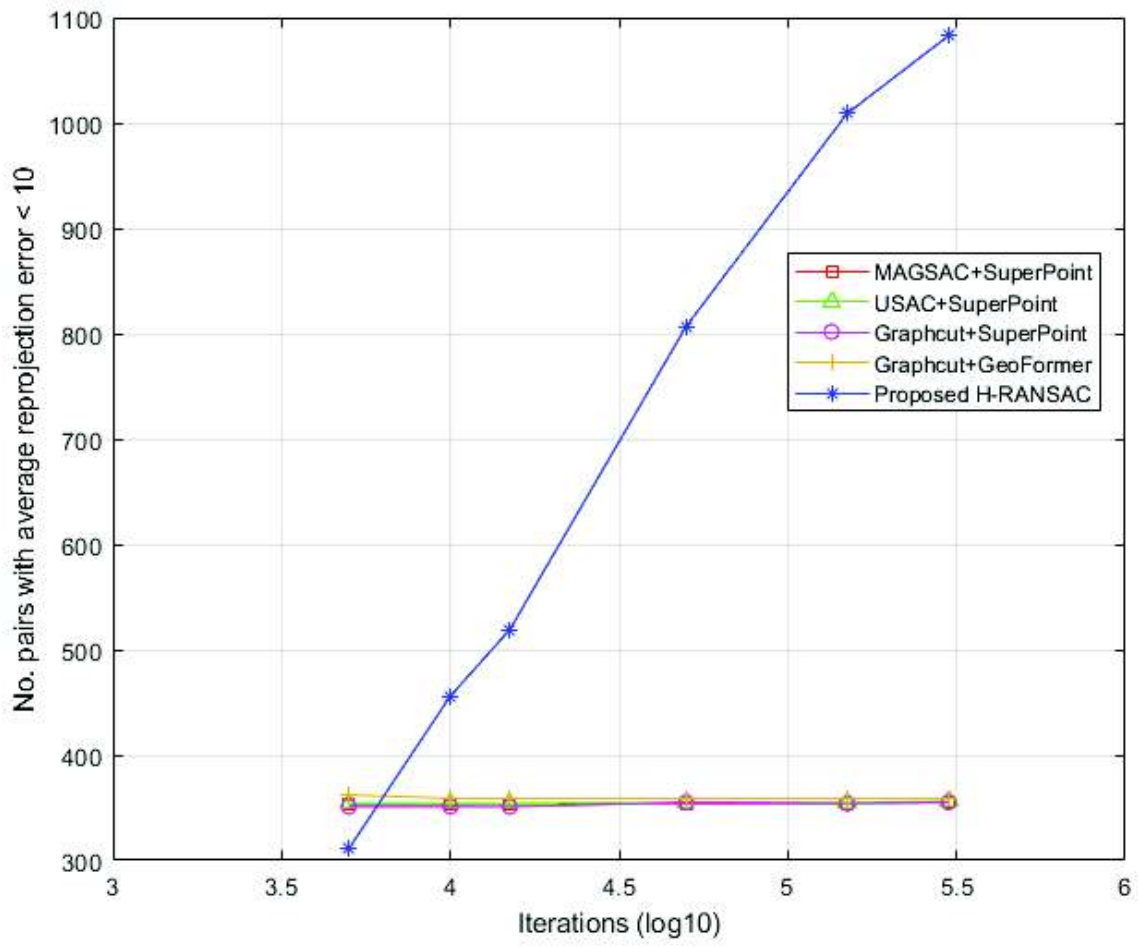

**FIGURE 17.** **Number of frame pairs aligned for a fixed number of iterations.**

The behavior of the proposed $H$-RANSAC is assessed in more detail in Fig. 13a, where the number of frame pairs with recovered $H$ by the algorithm is presented as a function of the number of correctly and incorrectly identified ground truth point pairs. It is obvious that in the majority of the processed frames the proposed method did not select even one wrong point-pair, thus in all these frames the homography transform was successfully recovered.

The details of the results achieved by Superpoint+Graph-Cut are shown in Fig. 13b, in complete analogy to Fig. 13a for the proposed $H$-RANSAC. The number of frame pairs with less than 4 correct point pairs are dominating the plot, however these frames are not aligned correctly.

Fig. 14 depicts the number of frame pairs with average reprojection error in [0-5] pixels (left group of bars), (5-10] (middle group) and >10 pixels right group by each method. It is clear that the proposed $H$-RANSAC had the greatest frame pair number with reprojection error within [0-5] pixels, followed closely by GeoFormer, DKM and Superpoint, combined with MAGSAC, Graph-Cut or VSAC. $H$-RANSAC also managed to match many more frames with reprojection error in (5,10] pixels. Boxplots of the average reprojection error are given in Fig. 15 for all methods, however care has to be taken when interpreting this graph, since the total number of frame pairs where $H$ matrix was recovered by each method is not shown. For instance, Graph-Cut+ORB and VSAC+ORB achieved low reprojection error, however, it is averaged over only 308 frame pairs that the two methods managed to recover $H$.

Deep learning-based methods for the estimation of image homography use the corner displacements of the image, rather than the actual elements of the $H$ matrix. The displacement range is usually 25% of the image dimensions. In the proposed work, the homography is estimated between radically different views. Having determined the matrix $H$, the corner displacements of the transformed image can be easily computed and plotted in Fig. 16. Considering the dimensions of each image frame, the average displacement along the X-axis is approximately 150%, well beyond the capabilities of the deep learning-based approaches.

Considering numerical complexity, the proposed $H$-RANSAC is a variant of the generic RANSAC algorithm, thus a single iteration has essentially the same order of complexity with an iteration of any other RANSAC variant. The main extra computational burden is the calculation of $Q$ for each random 4-point permutation according to Eq.(5) and the $Q$ value for the post-hic criterion. However, these can be considered negligible compared to calculating the $H$ matrix and the reprojection error, which dominate the calculations in each iteration. Thus, we compare the numerical burden of the RANSAC-based methods in terms of the required number of iterations, which is a more reliable measure since it is implementation independent. Figure 17 shows the number of image pairs successfully matched (average reprojection error per image pair less than 10 pixels) versus the maximum number of allowed iterations $n_{iter}$, for the proposed method and the best performing methods in comparison. A typical execution time for $10^5$ iterations was less than 5 seconds, in a typical laptop (i5-11$^{th}$ gen, 8GB RAM) using MATLAB 2021a. The proposed algorithm greatly benefits when increasing $n_{iter}$, whereas the rest of the methods seem unable to exploit the increased iterations.

## IV. CONCLUSION

The proposed $H$-RANSAC recovers the homography matrix between two images, using one set of unpaired points from each image, without local features, optionally assigned into two classes. A robust geometric criterion rejects implausible point selection before each iteration of RANSAC, based on the type of the quadrilaterals formed by random point pair selection (convex or concave and (non)-self-intersecting). Also, a similar post-hoc criterion that rejects implausible homography transformations is included at the end of each iteration. These criteria are similar to cheiral constraints but are less strict and more efficient, considering the fraction of 4-point selections.

The expected maximum iterations of $H$-RANSAC are derived for different probabilities of success, according to the number of points per image and per class, and the percentage of outliers, with and without considering geometric constraints.

The proposed methodology has been applied to a demanding dataset that combines 12 views of a football stadium per timestamp, many of them radically different to each other with respect to camera position, viewing vector and zoom. The available human annotations provide the means to assess the performance of the proposed algorithm. Several other state-of-the-art methods have also been applied to this dataset, combining different point selection algorithms (classic and deep learning-based) with established RANSAC variants.

It should be mentioned that the ground truth homography matrix $H$ between the reference frame and the rest of simultaneous frames is not known a priori. Furthermore, tt is well known that in the case of affine transformations, or in the case of Homography image stitching for very distant objects, accuracy metrics are based on similarity between reference and transformed image (such as RMSE, PSNR, SSIM etc.). In this specific case however, where we used homography to match planar points at finite distance from the cameras, only the pixels that lie on the same plane (i.e. the football field) should be considered when evaluating similarity criteria between homography-transformed images. The grandstand (which accommodates spectators), the goals, and the players and referees are all at an elevated level with respect to the planar field. Thus, the homography-transformed images of these objects will not match. This is evident for instance in Figures 7, 9, 11, where the players' shoes match between images, but the rest of the body is correctly mapped to different image pixels. Consequently, image similarity criteria cannot be applied to quantify the quality of homographic transform. Therefore, we resorted to the reprojection error between point pairs, which is usually the main error metric, in such cases [48], [49].

Results demonstrate that our dataset is very demanding, containing frames with extreme camera pose and zoom factor change. Therefore, out of 2312 frame pairs (of reference frames at any time stamp and any other camera frame with sufficient number of players according to Eq. (1)), the proposed method managed to recover $H$ for 1331 image pairs, of which, 939 aligned correctly. The best of the other methods (GeoFormer, DKM and SuperPoint, combined with MAGSAC++) recovered $H$ from significantly less frame pairs, even fewer of which aligned correctly. Furthermore, the deep learning method of Superpoint clearly outperforms the classic SIFT and ORB, for all three tested RANSAC variants.

We believe that the main advantage of the proposed algorithm is the dynamic construction of pairs from featureless points on the planar image parts, in contrast to other methods that rely on pre-determined candidate pairs of points, which may appear locally similar but will not match under homography transform. Thus, we may conclude that in cases with few tens of available points per image, without local features to form point pairs, the proposed $H$-RANSAC is the method of choice for homography recovery, especially for frame pairs with extreme geometric transformations.

In our task with few tens of featureless points per image, clustered into two distinct classes, the proposed $H$-RANSAC required on average $10^4 - 10^5$ iterations. Future work will concentrate on reducing further the number of iterations, or implementing efficient parallelization.

## REFERENCES

[1] M. A. Fischler and R. C. Bolles, "Random sample consensus: A paradigm for model fitting with applications to image analysis and automated cartography," *Commun. ACM*, pp. 726–740, 1987.

[2] G. Bradski, "The OpenCV library," *Dr. Dobb's J., Softw. Tools Prof. Programmer*, vol. 25, no. 11, pp. 120–125, 2000.

[3] P. H. S. Torr and A. Zisserman, "MLESAC: A new robust estimator with application to estimating image geometry," *Comput. Vis. Image Understand.*, vol. 78, no. 1, pp. 138–156, Apr. 2000.

[4] R. Raguram, O. Chum, M. Pollefeys, J. Matas, and J.-M. Frahm, "USAC: A universal framework for random sample consensus," *IEEE Trans. Pattern Anal. Mach. Intell.*, vol. 35, no. 8, pp. 2022–2038, Aug. 2013.

[5] O. Chum and J. Matas, "Matching with PROSAC—Progressive sample consensus," in *Proc. IEEE Comput. Soc. Conf. Comput. Vis. Pattern Recognit. (CVPR)*, vol. 1, Jun. 2005, pp. 220–226.

[6] D. Barath, J. Noskova, M. Ivashechkin, and J. Matas, "MAGSAC++, a fast, reliable and accurate robust estimator," in *Proc. IEEE/CVF Conf. Comput. Vis. Pattern Recognit. (CVPR)*, Jun. 2020, pp. 1301–1309.

[7] M. Ivashechkin, D. Barath, and J. Matas, "VSAC: Efficient and accurate estimator for h and f," in *Proc. IEEE/CVF Int. Conf. Comput. Vis. (ICCV)*, Oct. 2021, pp. 15223–15232.

[8] D. Barath and J. Matas, "Graph-cut RANSAC: Local optimization on spatially coherent structures," *IEEE Trans. Pattern Anal. Mach. Intell.*, vol. 44, no. 9, pp. 4961–4974, Sep. 2022.

[9] K. Ni, H. Jin, and F. Dellaert, "GroupSAC: Efficient consensus in the presence of groupings," in *Proc. IEEE 12th Int. Conf. Comput. Vis.*, Sep. 2009, pp. 2193–2200.

[10] O. Chum, T. Werner, and J. Matas, "Two-view geometry estimation unaffected by a dominant plane," in *Proc. IEEE Comput. Soc. Conf. Comput. Vis. Pattern Recognit. (CVPR)*, vol. 1, Jun. 2005, pp. 772–779.

[11] B. J. Tordoff and D. W. Murray, "Guided-MLESAC: Faster image transform estimation by using matching priors," *IEEE Trans. Pattern Anal. Mach. Intell.*, vol. 27, no. 10, pp. 1523–1535, Oct. 2005.

[12] D. DeTone, T. Malisiewicz, and A. Rabinovich, "SuperPoint: Self-supervised interest point detection and description," in *Proc. IEEE/CVF Conf. Comput. Vis. Pattern Recognit. Workshops (CVPRW)*, Jun. 2018, pp. 224–236.

[13] P.-E. Sarlin, D. DeTone, T. Malisiewicz, and A. Rabinovich, "SuperGlue: Learning feature matching with graph neural networks," in *Proc. IEEE/CVF Conf. Comput. Vis. Pattern Recognit. (CVPR)*, Jun. 2020, pp. 4937–4946.

[14] P. Mavrogiannis and I. Maglogiannis, "Amateur football analytics using computer vision," *Neural Comput. Appl.*, vol. 34, no. 22, pp. 19639–19654, Nov. 2022.

[15] B. D. Lucas and T. Kanade, "An iterative image registration technique with an application to stereo vision," in *Proc. 7th Int. joint Conf. Artif. Intell.*, 1981, pp. 674–679.

[16] G. Shi, X. Xu, and Y. Dai, "SIFT feature point matching based on improved RANSAC algorithm," in *Proc. 5th Int. Conf. Intell. Human-Mach. Syst. Cybern.*, vol. 1, Aug. 2013, pp. 474–477.

[17] H. Bay, T. Tuytelaars, and L. V. Gool, "SURF: Speeded up robust features," in *Proc. Eur. Conf. Comput. Vis.*, 2006, pp. 404–417.

[18] E. Rublee, V. Rabaud, K. Konolige, and G. Bradski, "ORB: An efficient alternative to SIFT or SURF," in *Proc. Int. Conf. Comput. Vis.*, Nov. 2011, pp. 2564–2571.

[19] W.-Y. Lin, M.-M. Cheng, J. Lu, H. Yang, N. Minh, and P. H. S. Torr, "Bilateral functions for global motion modeling," in *Proc. 13th Eur. Conf. Comput. Vis. (ECCV)*, Zurich, Switzerland, 2014, pp. 341–356.

[20] Z. Hossein-nejad and M. Nasri, "Image registration based on SIFT features and adaptive RANSAC transform," in *Proc. Int. Conf. Commun. Signal Process. (ICCSP)*, Apr. 2016, pp. 1087–1091.

[21] S. Liu, J. Chen, C.-H. Chang, and Y. Ai, "A new accurate and fast homography computation algorithm for sports and traffic video analysis," *IEEE Trans. Circuits Syst. Video Technol.*, vol. 28, no. 10, pp. 2993–3006, Oct. 2018.

[22] D. E. Goldberg, *Genetic Algorithms*. London, Pearson Education India, 2013.

[23] S. Liu, H. Wang, Y. Wei, and C. Pan, "BB-homography: Joint binary features and bipartite graph matching for homography estimation," *IEEE Trans. Circuits Syst. Video Technol.*, vol. 25, no. 2, pp. 239–250, Feb. 2015.

[24] J. Zhang, Y. Liu, G. Ding, B. Tang, and Y. Chen, "Adaptive decomposition and extraction network of individual fingerprint features for specific emitter identification," *IEEE Trans. Inf. Forensics Security*, vol. 19, pp. 8515–8528, 2024.

[25] J. Zhang, S. Huang, J. Liu, X. Zhu, and F. Xu, "PYRF-PCR: A robust three-stage 3D point cloud registration for outdoor scene," *IEEE Trans. Intell. Vehicles*, vol. 9, no. 1, pp. 1270–1281, Jan. 2024.

[26] D. DeTone, T. Malisiewicz, and A. Rabinovich, "Deep image homography estimation," 2016, *arXiv:1606.03798*.

[27] L. Nie, C. Lin, K. Liao, S. Liu, and Y. Zhao, "Depth-aware multi-grid deep homography estimation with contextual correlation," *IEEE Trans. Circuits Syst. Video Technol.*, vol. 32, no. 7, pp. 4460–4472, Jul. 2022.

[28] T. Nguyen, S. W. Chen, S. S. Shivakumar, C. J. Taylor, and V. Kumar, "Unsupervised deep homography: A fast and robust homography estimation model," *IEEE Robot. Autom. Lett.*, vol. 3, no. 3, pp. 2346–2353, Jul. 2018.

[29] T. Liao, X. Zhang, Y. Xu, Z. Shi, and G. Xiao, "SGA-Net: A sparse graph attention network for two-view correspondence learning," *IEEE Trans. Circuits Syst. Video Technol.*, vol. 33, no. 12, pp. 7578–7590, Dec. 2023.

[30] Y. Tian, X. Yu, B. Fan, F. Wu, H. Heijnen, and V. Balntas, "SOSNet: Second order similarity regularization for local descriptor learning," in *Proc. IEEE/CVF Conf. Comput. Vis. Pattern Recognit. (CVPR)*, Jun. 2019, pp. 11008–11017.

[31] S. Khan, M. Nawaz, X. Guoxia, and H. Yan, "Image correspondence with CUR decomposition-based graph completion and matching," *IEEE Trans. Circuits Syst. Video Technol.*, vol. 30, no. 9, pp. 3054–3067, Sep. 2020.

[32] J.-S. Park, H.-E. Kim, H.-Y. Kim, J. Lee, and L.-S. Kim, "A vision processor with a unified interest-point detection and matching hardware for accelerating a stereo-matching algorithm," *IEEE Trans. Circuits Syst. Video Technol.*, vol. 26, no. 12, pp. 2328–2343, 2015.

[33] M. Hong, Y. Lu, N. Ye, C. Lin, Q. Zhao, and S. Liu, "Unsupervised homography estimation with coplanarity-aware GAN," in *Proc. IEEE/CVF Conf. Comput. Vis. Pattern Recognit. (CVPR)*, Jun. 2022, pp. 17642–17651.

[34] S.-Y. Cao, J. Hu, Z. Sheng, and H. Shen, "Iterative deep homography estimation," in *Proc. IEEE/CVF Conf. Comput. Vis. Pattern Recognit.*, Jun. 2022, pp. 1869–1878.

[35] Q. Zhou and X. Li, "Deep homography estimation and its application to wall maps of wall-climbing robots," *Appl. Sci.*, vol. 9, no. 14, p. 2908, Jul. 2019.

[36] C. Shao, C. Zhang, Z. Fang, and G. Yang, "A deep learning-based semantic filter for RANSAC-based fundamental matrix calculation and the ORB-SLAM system," *IEEE Access*, vol. 8, pp. 3212–3223, 2020.

[37] G. Wang, X. Sun, Y. Shang, Z. Wang, Z. Shi, and Q. Yu, "Two-view geometry estimation using RANSAC with locality preserving constraint," *IEEE Access*, vol. 8, pp. 7267–7279, 2020.

[38] J. Zheng, W. Peng, Y. Wang, and B. Zhai, "Accelerated RANSAC for accurate image registration in aerial video surveillance," *IEEE Access*, vol. 9, pp. 36775–36790, 2021.

[39] J. Liu and X. Li, "Geometrized transformer for self-supervised homography estimation," in *Proc. IEEE/CVF Int. Conf. Comput. Vis. (ICCV)*, Oct. 2023, pp. 9556–9565.

[40] H. Zhu, S.-Y. Cao, J. Hu, S. Zuo, B. Yu, J. Ying, J. Li, and H.-L. Shen, "MCNet: Rethinking the core ingredients for accurate and efficient homography estimation," in *Proc. IEEE/CVF Conf. Comput. Vis. Pattern Recognit. (CVPR)*, Jun. 2024, pp. 25932–25941.

[41] X. Deng, E. Liu, C. Gao, S. Li, S. Gu, and M. Xu, "CrossHomo: Cross-modality and cross-resolution homography estimation," *IEEE Trans. Pattern Anal. Mach. Intell.*, vol. 46, no. 8, pp. 5725–5742, Aug. 2024.

[42] S.-Y. Cao, R. Zhang, L. Luo, B. Yu, Z. Sheng, J. Li, and H.-L. Shen, "Recurrent homography estimation using homography-guided image warping and focus transformer," in *Proc. IEEE/CVF Conf. Comput. Vis. Pattern Recognit. (CVPR)*, Jun. 2023, pp. 9833–9842.

[43] J. Edstedt, I. Athanasiadis, M. Wadenbäck, and M. Felsberg, "DKM: Dense kernelized feature matching for geometry estimation," in *Proc. IEEE/CVF Conf. Comput. Vis. Pattern Recognit. (CVPR)*, Jun. 2023, pp. 17765–17775.

[44] G. Nakano, "Algebraic constraint for preserving convexity of planar homography," in *Proc. Int. Conf. 3D Vis. (3DV)*, Dec. 2021, pp. 126–135.

[45] H. Zhu, X. Wen, F. Zhang, X. Wang, and G. Wang, "Homography estimation based on order-preserving constraint and similarity measurement," *IEEE Access*, vol. 6, pp. 28680–28690, 2018.

[46] D. E. Goldberg, *Genetic Algorithms in Search, Optimization and Machine Learning*, 1st ed., Reading, MA, USA: Addison-Wesley, 1989.

[47] G. Jocher, C. Liu, A. Hogan, L. Y. Changyu, P. Rai, and T. Sullivan, "Ultralytics/YOLOv5: Initial release," Zenodo, Tech. Rep., 2020.

[48] Y. Luo, X. Wang, Y. Liao, Q. Fu, C. Shu, Y. Wu, and Y. He, "A review of homography estimation: Advances and challenges," *Electronics*, vol. 12, no. 24, p. 4977, Dec. 2023.

[49] S. K. Grzechnik, "Devices, methods, and apparatuses for homography evaluation involving a mobile device," U.S. Patent 9 342 886, May 17, 2016.

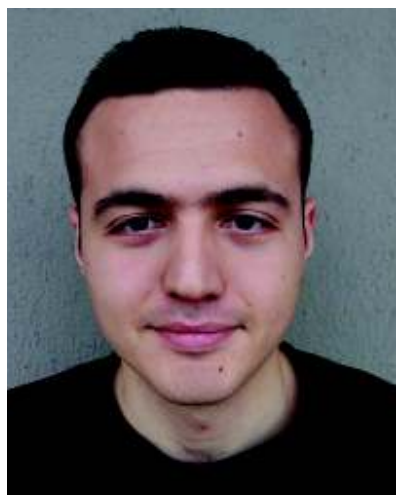

**GEORGE NOUSIAS** received the degree from the Department of Computer Science and Biomedical Informatics, University of Thessaly, Greece, and the M.Sc. degree in artificial intelligence from the Department of Computer Science, Aristotle University of Thessaloniki. He is currently pursuing the Ph.D. degree with the Department of Computer Science and Biomedical Informatics, University of Thessaly.

His research interests include computer vision, machine learning, and the analysis of biomedical signals, images, and videos.

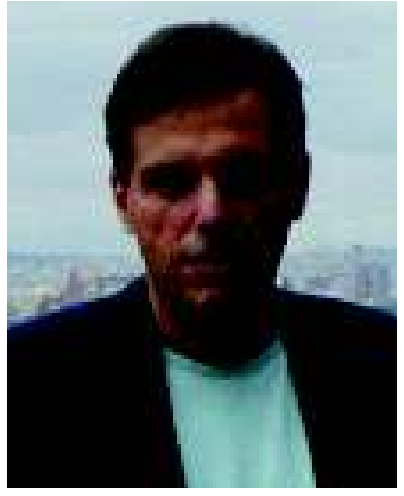

**KONSTANTINOS K. DELIBASIS** received the degree from the Department of Physics, University of Athens, Greece, and the M.Sc. degree in medical physics and the Ph.D. degree in medical image analysis from the Department of Biomedical Physics and Bio-Engineering, University of Aberdeen, U.K.

He is currently a Professor with the Department of Computer Science and Biomedical Informatics, University of Thessaly. He has co-authored more than 70 scientific journal articles. He has taught more than ten different undergraduate and postgraduate courses. His research interests include the processing and analysis of biomedical signals, images, and video, and computer simulations of physical and biomedical processes.

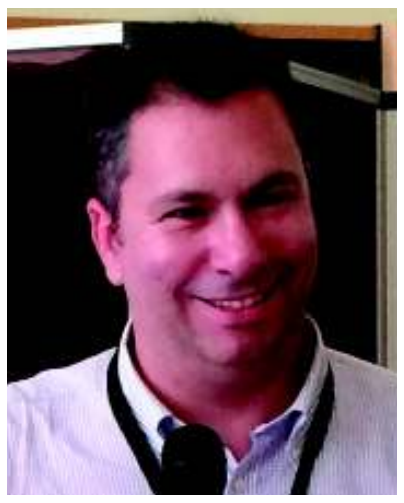

**ILIAS G. MAGLOGIANNIS** (Senior Member, IEEE) received the Diploma degree in electrical and computer engineering and the Ph.D. degree in biomedical engineering and medical informatics from the National Technical University of Athens, Athens, Greece.

He is currently an Associate Professor and the Director of the Computational Biomedicine Laboratory, Department of Digital Systems, University of Piraeus, Piraeus, Greece. His published scientific work includes five books, more than 120 journal articles, and over 180 international conference papers. He has received approximately 10 000 citations for his research work. His current research interests include biomedical informatics, computer vision, multimedia processing, and pervasive healthcare systems.